%% file: main_camera_ready.tex
\documentclass[nohyperref]{article}

\usepackage{microtype}
\usepackage{graphicx}
\usepackage{subfigure}
\usepackage{booktabs} 

\usepackage{hyperref}

\usepackage[accepted]{icml2022}

\usepackage{amsmath}
\usepackage{amssymb}
\usepackage{mathtools}
\usepackage{amsthm}

\usepackage[capitalize,noabbrev]{cleveref}

\theoremstyle{plain}

\theoremstyle{definition}

\theoremstyle{remark}

\usepackage[textsize=tiny]{todonotes}

\icmltitlerunning{Learning to Estimate and Refine Fluid Motion with Physical Dynamics}

\begin{document}

\twocolumn[
\icmltitle{Learning to Estimate and Refine Fluid Motion with Physical Dynamics}



\icmlsetsymbol{equal}{*}

\begin{icmlauthorlist}
\icmlauthor{Mingrui Zhang}{ese}
\icmlauthor{Jianhong Wang}{eee}
\icmlauthor{James Tlhomole}{ese}
\icmlauthor{Matthew D. Piggott}{ese}
\end{icmlauthorlist}

\icmlaffiliation{ese}{Department of Earth Science and Engineering, Imperial College London, UK}
\icmlaffiliation{eee}{Department of Electrical and Electronic Engineering, Imperial College London, UK}

\icmlcorrespondingauthor{Mingrui Zhang}{mingrui.zhang18@imperial.ac.uk}

\icmlkeywords{Machine Learning, ICML}

\vskip 0.3in
]



\printAffiliationsAndNotice{}

\begin{abstract}
Extracting information on fluid motion directly from images is challenging. Fluid flow represents a complex dynamic system governed by the Navier-Stokes equations. General optical flow methods are typically designed for rigid body motion, and thus struggle if applied to fluid motion estimation directly. Further, optical flow methods only focus on two consecutive frames without utilising historical temporal information, while the fluid motion (velocity field) can be considered a continuous trajectory constrained by time-dependent partial differential equations (PDEs). This discrepancy has the potential to induce physically inconsistent estimations. Here we propose an unsupervised learning based prediction-correction scheme for fluid flow estimation. An estimate is first given by a PDE-constrained optical flow predictor, which is then refined by a physical based corrector. The proposed approach outperforms optical flow methods and shows competitive results compared to existing supervised learning based methods on a benchmark dataset. Furthermore, the proposed approach can generalize to complex real-world fluid scenarios where ground truth information is effectively unknowable. Finally, experiments demonstrate that the physical corrector can refine flow estimates by mimicking the operator splitting method commonly utilised in fluid dynamical simulation.
\end{abstract}

\input{introduction}
\input{background}
\input{method}
\input{experiment}
\input{related_work}
\input{conclusion}


\bibliography{references}
\bibliographystyle{icml2022}

\input{appendix}


\end{document}

%% file: introduction.tex
\section{Introduction}

Fluid flow motion estimation is a topic of interest for many science and engineering fields, including geophysics, oceanology, biology, and environmental engineering. Measuring fluid motion and understanding the underlying dynamics are crucial for exploring complex fluid phenomena in these fields.

One natural way to understand and analyze fluid motion is via visual observation. However, there are generally no strong visible patterns in transparent fluid flow such as water and air. Therefore, visual markers of some description are commonly introduced to allow the optical measurement of the motion. One effective method is to inject these markers into the fluid and record their motion with one or multiple high-speed cameras. By comparing the resulting flow images at different time levels, velocity field information can be extracted. 
Based on this idea, one of most prominent techniques for fluid motion estimation in experimental fluid mechanics is Particle Image Velocimetry (PIV) \cite{RAdrian}. Traditionally, PIV can be regarded as an optical flow estimation problem \citep{Ruhnau_of_piv} and tackled using variational optical flow methods \citep{Heitz_2010}. 

With the success of deep learning in optical flow estimation, 
these methods have also been adopted to solve the corresponding PIV problem. However, pure dense optical flow methods assume brightness constancy and flow smoothness, while the visible tracers in the fluid are driven by fluid dynamics.
The missing dynamical information in the estimation model may induce physically implausible results and temporal inconsistency, which would typically be crucial for useful flow diagnosis in rigorous science and engineering applications. In this work we seek to bridge the gap between optical estimation and the governing fluids-based physical models for motion estimation problem. Specifically, a novel unsupervised learning framework is proposed in this paper. The framework is designed as a prediction-correction scheme, which consists of an optical flow based fluid motion predictor and a physical corrector. 

The estimation process is inspired by Chorin’s projection method, an operator splitting method \citep{chorin, operator_spliting}, often used in numerical fluid simulation. The operator splitting method separates the original PDE system into two parts over a time step, separately computes the solution to each part, and then combines the two separate solutions to form a solution to the original system. However, these kinds of methods are limited to physical models, and thus cannot incorporate optical fluid observations, and this cannot be used for fluid motion estimation problem directly. Therefore, here we consider an approach based upon a generalization of the projection method to a generic operator splitting scheme, which incorporates both physical knowledge and fluid observations. In this scheme, the motion predictor first outputs an estimated flow field. This is then refined using a physical based corrector.

For the motion predictor, we note that the Euler-Lagrange equation corresponding to the unsupervised learning formulation is related to the Stokes equation in fluid dynamics. The Stokes equation is a simplified version of the full Navier-Stokes equation, which indicates that the estimation result can in some sense be considered a component of the full equation; thus motivating its inclusion in an operator splitting like scheme. For the physical corrector, the approach taken uses both the velocity field from the previous time level and the current estimate as inputs and is designed to enforce physical consistency and the divergence-free constraint in one shot. We test our resulting estimations methods on both synthetic and real world datasets. The experiments indicate that the proposed unsupervised method can output competitive results on synthetic dataset compared to a supervised method. For the real world dataset without ground truth, the method can achieve reasonable estimation results as noted through benchmarking against those obtained using state-of-the-art open-source fluid motion estimation software. Code is available at: \url{https://github.com/erizmr/Learn-to-Estimate-Fluid-Motion}.

%% file: background.tex
\section{Background}
\subsection{Fluid Flow}
Fluid dynamics is typically modelled using the Navier-Stokes (N-S) equations. There are different variations of the N-S  equations depending on the nature of the fluid or its flow. Here we briefly introduce the system that we consider in this work.

\textbf{Incompressible Fluid Flow.} In this work, we only consider incompressible flow, which can be modelled using the Navier-Stokes momentum equation 
\begin{equation}
    \begin{aligned}
    \frac{\partial \bf{u}}{\partial t} + {\bf{u}} \cdot \nabla {\bf{u}} =-\frac{1}{\rho}\nabla p + \nu \nabla^2 \bf {u} + {\bf{f}},
    \end{aligned}
    \label{eq:NS-incompressible}
\end{equation}
where ${\bf u}$ is the velocity, $p$ denotes the pressure field (scalar field), $\rho$ is the fluid density (assumed to be constant), $\nu$ is the kinematic viscosity (also assumed to be constant) and ${\bf f}$ is the summation of any external forces applied to the fluid body. To represent incompressibility, a divergence-free constraint on the velocity vector field: $\nabla \cdot {\bf{u}} = 0$, should be satisfied.

\textbf{Transport Equation and Warping.}\label{paragraph:transport}
A scalar field $I$ transported in and by the fluid can be described by the advection-diffusion equation
\begin{equation}
    \begin{aligned}
    \frac{\partial I}{\partial t} + \nabla \cdot (I{\bf{u}}) = D\nabla^2 I,
    \end{aligned}
    \label{eq:convection}
\end{equation}
where $D$ is the diffusion coefficient. Given a scalar field $I$ transported in incompressible flow, and assuming it is conserved in the region of interest, i.e., no source terms appear, and that the diffusion coefficient $D=0$, then with the divergence-free condition, Equation \eqref{eq:convection} can be simplified to
\begin{equation}
    \begin{aligned}
    \frac{\partial I}{\partial t} + {\bf{u}} \cdot \nabla {I} = 0.
    \end{aligned}
    \label{eq:bca}
\end{equation}
Note that Equation \eqref{eq:bca} is consistent with the brightness constancy assumption in Horn and Schunck's optical flow approach \cite{Horn_optical_flow}. Therefore, optical flow can be regarded as a special case of fluid flow, i.e., visible markers move in a purely advective manner in fluid flow.

For Equation \eqref{eq:convection}, it can be shown \citep{dl_physics_process} that given any initial condition $I_0$, there exists a unique solution $I({\bf x}, t)$ which can be computed via a convolution between a Gaussian kernel and the initial condition $I_0$ ( shown in Appendix \ref{warping_proof}). Therefore, we can obtain the warping scheme below by discretizing the solution. Using a previous time level scalar field $I_t$ as the initial condition, we can compute the image at the next time level as
\begin{equation}
    \begin{aligned}
     \hat{I}_{t+1}({\bf x})  = \sum_{{\bf y} \in \Omega} k({\bf x}-{\bf u}, {\bf y})I_t({\bf y}),
    \end{aligned}
    \label{eq:warping_scheme_2}
\end{equation}
where $k({\bf x} -{\bf u}, {\bf y}) = \frac{1}{4\pi D\delta t}e^{-\frac{1}{4D\delta t}\left\| {\bf x} -{\bf u} - {\bf y} \right\|^2}$, $\delta t$ is the time interval between time levels $t$ and $t+1$. Equation \eqref{eq:warping_scheme_2} shares similar ideas with the Spatial Transformer Network \citep{warp}, where the $k(\cdot)$ is a sampling kernel. By using this warping scheme, we can approximate the solution of scalar fields such as the vorticity and the concentration of the marker in the fluid, which will be utilised in training the corrector and predictor in this work.

%% file: method.tex
\section{Learning to Estimate and Refine Fluid Motion}
\begin{figure}
\centering
\includegraphics[width=1.0\columnwidth]{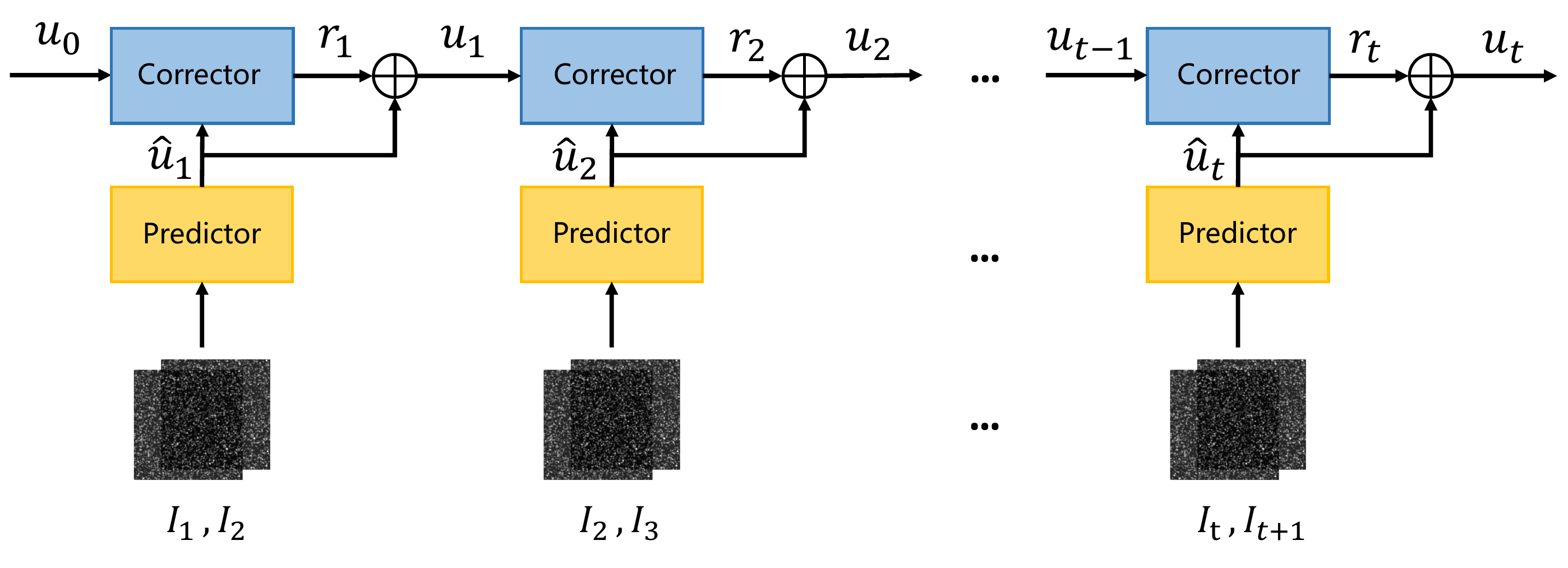}
\caption{The pipeline of the fluid motion estimation via a prediction-correction based scheme. Given an image pair $I_t, I_{t-1}$ at the current time level $t$, a predictor first outputs an estimated velocity field $\hat{\bf u}_t$. Then a corrector computes a refinement ${\bf r}_t$ by taking both the velocity at the previous time level, i.e. ${\bf u}_{t-1}$ and the current estimate $\hat{\bf u}_t$ as inputs. By adding the current estimate and the refinement, the corrected ${\bf u}_t$ is computed.}
\end{figure}

\textbf{Problem Statement.} Given a sequence of consecutive fluid observation images ${\bf I} = (I_1, I_2, I_3, ...,I_T) \in \mathbb{R}^{T \times c \times h  \times w}$ as input, where $T$, $c$, $h$, $w$ are number of total time steps, number of image channels, height and width of images. Our goal is to estimate the dense forward flow field (displacement field) for each pair of images, from $\{I_1, I_2\}$ to $\{I_{T-1}, I_T\}$, which is denoted as ${\bf{u}} = ({\bf u}_1,...,{\bf u}_{T-1}) \in \mathbb{R}^{(T-1) \times 2 \times h  \times w}$. In addition, the fluid flow field is assumed to be governed by the incompressible Navier-Stokes Equation \eqref{eq:NS-incompressible}.

We propose an unsupervised learning based prediction-correction scheme for the fluid motion estimation problem. Given a time step $t$, an optical flow based predictor $\mathcal{P}$ provides an estimated flow field ${ \hat{\bf u}_t}$. Then a physical corrector takes both ${ \hat{\bf u}_t}$ and ${\bf u}_{t-1}$ (the flow field from the previous time level) as inputs, and outputs a refinement in the form of a corrected flow field ${\bf u}_t$: 
\begin{equation}
    \begin{aligned}
     {\bf u}_t = \mathcal{ P}(I_{t-1}, I_{t}) + \mathcal{C}({\bf u}_{t-1}, \hat{\bf u}_{t}).
    \end{aligned}
    \label{eq:p_c_scheme}
\end{equation}
\subsection{Fluid Motion Predictor} The fluid motion estimator is designed based on a variational optical flow approach constrained by the Stokes equation. The unsupervised learning formulation is given by
\begin{equation}
    \begin{aligned}
    \min_{{\bf{u}}} \int_\Omega  \underbrace{{ \left( \frac{\partial I}{\partial t} + {\bf u} \cdot \nabla I \right)}}_{Data \,\, term} +  \underbrace{\mu |\nabla {\bf u}|^2}
    _{Smoothness} + \underbrace{p\nabla \cdot {\bf u} }_{Divergence}  \,d\bf{x}. 
    \end{aligned}
    \label{eq:total_energy_spatial}
\end{equation}
The formula consists of three parts: the data term, the smoothness and the divergence regularizers. The data term is modelled by the brightness constancy assumption \eqref{eq:bca}. For the smoothness and divergence regularizer, these can be identified with the diffusion (or viscous) and pressure gradient terms in the fluid dynamical equations as shown below.

It can be shown (in Appendix \ref{el_proof}) that the Euler-Lagrange equation of formulation \eqref{eq:total_energy_spatial} is:
\begin{equation}
    \begin{aligned}
     - \mu \nabla^2 {\bf {u}} + \nabla p = \nabla I,
    \end{aligned}
    \label{eq:stokes_equation_predictor}
\end{equation}
which has the same form as the Stokes equation in fluid dynamics. It is is an approximation/simplification to the Navier-Stokes momentum equation obtained by omitting the inertial component. The $\nabla I$ can be interpreted as an external forcing term, which can be determined by optical observations. The forward process of a neural network model has a potential mathematical equivalence to the temporal evolution of a dynamic system \citep{E_dynamic}. Thus, the training process of the predictor can be regarded as finding the optimal control forces applied on the fluid dynamical system so as to minimize the energy \eqref{eq:total_energy_spatial}. It also suggests that the inference process is solving for a velocity field that is related to the Stokes equation given the optical observations.

\subsection{Physical Corrector} 
\textbf{Operator Splitting Scheme.} The idea of the corrector is motivated by the operator splitting approach common in the numerical solution of PDEs. The approach separates the original equation into two or more parts and computes the solution to each part separately. The separate solutions are then combined to form a solution to the original equation.
In incompressible fluid simulation, the flow is often solved using a operator splitting based approach that is often referred to as Chorin’s projection method \cite{chorin}. The idea is first to compute a tentative velocity ${\bf u}_t^*$ by neglecting the pressure in the Navier-Stokes momentum equation and then to project the velocity onto the space of divergence free vector fields.

However, Chorin's projection method does not incorporate fluid observations into the scheme, and thus can not be adopted to fluid motion estimation problems directly. In addition, it often involves the solution of a Poisson equation in order to enforce the divergence-free condition, which is expensive compared to conducting neural network based inference. 
In this work, motivated by Chorin's projection method (a detailed introduction to which is given in Appendix \ref{chorin_method}), we generalise to a generic operator splitting scheme of the form 
\begin{equation}
    \begin{aligned}
      \frac{{\bf u}_t^* - {\bf u}_{t-1}}{\Delta t} = - {\bf u}_{t-1} \cdot \nabla{{\bf u}_{t-1}} + \nu \nabla^2{{\bf u}_{t-1}},
    \end{aligned}
    \label{eq:chorin_projection_predictor_1}
\end{equation}
\begin{equation}
    \begin{aligned}
      \frac{{\bf u}_t - {\bf u}_t^*}{\Delta t} = - \frac{1}{\rho} \nabla{p_{t}}.
    \end{aligned}
    \label{eq:chorin_projection_predictor_2}
\end{equation}
By adding up the two splitting parts i.e., the Equation \eqref{eq:chorin_projection_predictor_1} and Equation \eqref{eq:chorin_projection_predictor_2}, we can recover the incompressible Navier-Stokes equation with temporal discretization
\begin{equation}
    \begin{aligned}
      \frac{{\bf u}_t - {\bf u}_{t-1}}{\Delta t} = - \frac{1}{\rho} \nabla{p_{t}} + \mathcal{R}_1.
    \end{aligned}
    \label{eq:n_s_predictor}
\end{equation}
For clarity, we use $\mathcal{R}_1$ to denote the advection and viscous terms, i.e., $\mathcal{R}_1 = - {\bf u}_{t-1} \cdot \nabla{{\bf u}_{t-1}} + \nu \nabla^2{{\bf u}_{t-1}}$. Since solving for pressure $p_t$ is expensive, we turn to compute the pressure gradient term $\frac{1}{\rho} \nabla{p_{t}}$ using the left hand side of Equation \eqref{eq:chorin_projection_predictor_2}. Although ${\bf u}_t$ is an unknown, we can approximate it using the tentative velocity ${\bf u}_t^*$ and velocity estimation $\hat{\bf u}_t$ from the predictor:   
\begin{equation}
    \begin{aligned}
     \Tilde{\bf u}_t = {\bf K}_t \odot \hat{\bf u}_t + (1-{\bf K}_t)\odot {\bf u}_t^*,
    \end{aligned}
    \label{eq:u_tilde}
\end{equation}
where $\Tilde{\bf u}_t$ is the approximation to ${\bf u}_t$, a weighted average of the estimated and the tentative velocity. ${\bf K}_t$ here is a factor that controls the trade-off between the optical estimate and the physical velocity based on an advection-diffusion step. This control factor can be interpreted as the Kalman gain: if ${\bf K}_t = 1$, the output only relies on the estimation; while if ${\bf K}_t = 0$, the current velocity is computed fully by the physical advection-diffusion step. This can be modelled in a similar manner to the gating mechanisms in recurrent neural networks. ${\bf K}_t = \sigma({\bf W}_e*\hat{\bf u}_t+ {\bf W}_p*{\bf u}_t^* + {\bf b}_k),$ where ${\bf W}_e$ and ${\bf W}_p$ are convolutions and ${\bf b}_k$ is the bias, and $\sigma$ is the sigmoid function.

Substituting Equation \eqref{eq:chorin_projection_predictor_2} and Equation \eqref{eq:u_tilde} into Equation \eqref{eq:n_s_predictor} yields
\begin{equation}
    \begin{aligned}
      \frac{{\bf u}_t - {\bf u}_{t-1}}{\Delta t} = \frac{\Tilde{\bf u}_t - {\bf u}_t^*}{\Delta t} + \mathcal{R}_1 + \mathcal{R}_2,
    \end{aligned}
    \label{eq:n_s_predictor_2}
\end{equation}
\begin{equation}
     \begin{aligned}
     \frac{{\bf u}_t - \Tilde{\bf u}_t}{\Delta t} &= \mathcal{R}_2.
     \end{aligned}
     \label{eq:residual_2}
\end{equation}
$\mathcal{R}_2$ is used to model the physical residual induced by the predictor and the neglection of the pressure gradient term. In other words, the proposed scheme aims to ``project'' the errors in the predictor and the velocity field's divergence in one shot.

\textbf{Dynamics Model.} By reformulating Equation \eqref{eq:residual_2} as ${\bf u}_t = \Tilde{\bf u}_t + \Delta t \mathcal{R}_2$, we can compute the corrected velocity ${\bf u}_t$. Note that $\Tilde{\bf u}_t$ can be computed from ${\bf u}_t^*$ and $\hat{\bf u}_t$, and the only unknown is then $\mathcal{R}_2$.
$\mathcal{R}_2$ is designed to compensate the missing dynamics induced by the predictor and in neglecting the pressure gradient term. It is assumed that the missing dynamics can be modelled by PDEs and learned from data \citep{pde_net}. 
We consider an expression of the form 
\begin{equation}
    \begin{aligned}
     {\bf \Gamma}({\bf \Psi}({\bf x}, t)) = \sum_{i,j:i+j<q}c_{i,j}\frac{\partial^{i+j}{\bf \Psi}}{\partial x^i \partial y^j}({\bf x}, t),
    \end{aligned}
    \label{eq:model_derivatives}
\end{equation}
which combines spatial derivatives with coefficients $c_{i,j}$ up to a certain differential order $q$.
This is a generic linear combination of partial derivatives, which can be used as a basis to model a wide range of classical physical models, such as the advection-diffusion equation. Thus, $\mathcal{R}_2$ can be modelled using Equation \eqref{eq:model_derivatives}, wherein ${\bf \Psi}({\bf x}, t) \triangleq \hat{\bf u}_t(\textbf{x}, t) - {\bf u}_{t-1}$. Accordingly, we redefine the function ${\bf \Phi} ({\bf u}_{t-1}, \hat{\bf u}_{t}) \triangleq {\bf \Gamma}\left(\hat{\bf u}_t(\textbf{x}, t) - {\bf u}_{t-1}\right)$ for conciseness.

\textbf{Prediction-Correction Scheme.} To conclude, we can derive the prediction-correction scheme as:
\begin{equation}
    \begin{aligned}
      {\bf u}_t = \underbrace{{\bf K}_t \odot \hat{\bf u}_t}_{Predictor} + \underbrace{(1-{\bf K}_t)\odot {\bf u}_t^* + {\bf \Phi}({\bf u}_{t-1}, \hat{\bf u}_t)}_{Corrector},
    \end{aligned}
    \label{eq:pc_scheme_detail}
\end{equation}
where the predictor and corrector, denoted $\mathcal{P}(I_{t-1}, I_t)$ and $\mathcal{C}({\bf u}_{t-1}, \hat{\bf u}_t)$ respectively, are as mentioned in Equation \eqref{eq:p_c_scheme}. For the implementation we reformulate Equation \eqref{eq:pc_scheme_detail} as
\begin{equation}
    \begin{aligned}
      {\bf u}_t = {\bf u}_t^* + {\bf K}_t \odot (\hat{\bf u}_t-{\bf u}_t^*) + {\bf \Phi}({\bf u}_{t-1}, \hat{\bf u}_t).
    \end{aligned}
    \label{eq:pc_scheme_detail_2}
\end{equation}

\section{Implementation}
\begin{figure*}
\centering
\includegraphics[width=0.95\textwidth]{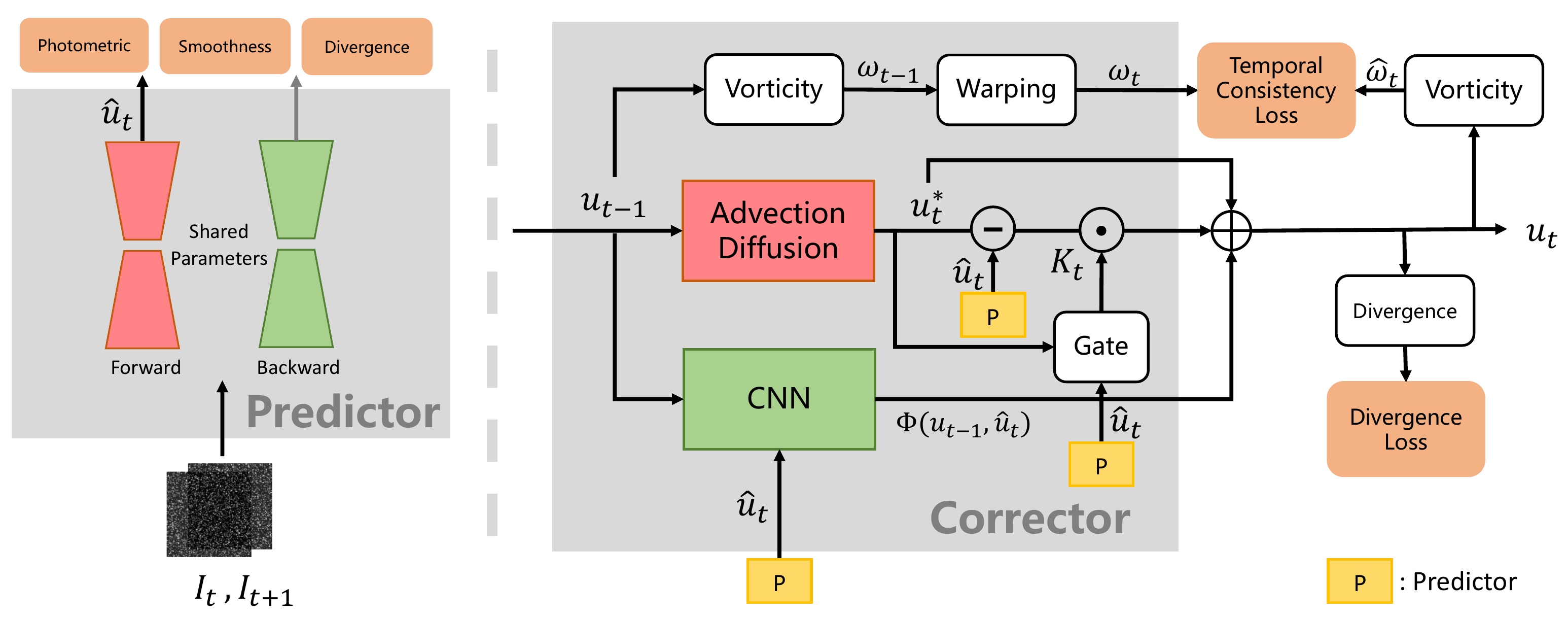}
\caption{The implementation of the predictor and the corrector described by Equation \eqref{eq:pc_scheme_detail_2}. For the predictor, the red and green blocks represent the network backbone, which is used to perform both forward and backward inference. For the corrector, the red block represents the fluid dynamics described by $\mathcal{R}_1$, which advects and diffuses the previous velocity to the current time level. The green block represents convolution layers which are used to compensate for missing dynamics described by Equation \eqref{eq:model_derivatives}. The gate models the control factor ${\bf K}_t$. The difference between the warped vorticity and the computed vorticity using the current velocity attempts to enforce temporal consistency. In addition, the divergence is also penalized during training.}
\label{fig:pipeline}
\end{figure*}

An overview of the predictor and corrector implementation is shown in  Figure \ref{fig:pipeline}. 
\subsection{Predictor} We use PWC-Net \citep{pwc} and LiteFlowNet \citep{twhui} as the backbone of the predictor. These two networks have an encoder-decoder like architecture and have been demonstrated to be successful for optical flow estimation problems. We re-train the network on synthetic fluid observation images in an unsupervised way. Due to the lack of annotated data for unsupervised learning, we take the bidirectional \citep{Meister_unflow} estimate into consideration to enrich the information used in the training loss. Therefore, two flow fields (the forward and backward flow) are defined respectively for each image pair as  ${\bf{u}}^f \equiv (u^f, v^f)^T$ and ${\bf{u}}^b \equiv (u^b, v^b)^T$.

\textbf{Training Loss.} As described in Equation \eqref{eq:total_energy_spatial}, the training loss of the predictor consists of three parts. The total loss is the weighted summation such that
\begin{equation}
    \begin{aligned}
    L_{P} = L_d + \lambda_s L_s + \lambda_{d} L_{div},
    \end{aligned}
    \label{eq:total_loss_definition}
\end{equation}
where $L_d$ denotes the data term, which is modelled by photometric loss; $L_s$ and $L_{div}$ are the spatial smoothness and divergence-free regularizers. $\lambda_s$ and $\lambda_d$ are the weights of the two regularizers respectively.

\textbf{Data Term.}
The data term is expressed in terms of the difference between warped and original input images, i.e., photometric loss. The bidirectional photometric loss is thus defined as the sum of these two parts:
\begin{align}
    L_d(I_1, I_2, {\bf{u}}^f, {\bf{u}}^b) = \sum_{{\bf{x}} \in P} \sigma \left(I_1({\bf{x}}) - \hat{I}_1({\bf x}) \right) \notag \\ + \sigma \left(I_2({\bf{x}}) - \hat{I}_2({\bf{x}}) \right),
\end{align}
where $\hat{I}_1({\bf x}) = I_2({\bf{x}}+ {\bf{u}}^f({\bf {x}}))$, $\hat{I}_2({\bf x}) = I_1({\bf{x}}+ {\bf{u}}^b({\bf {x}}))$ are the warped image for $I_1$ and $I_2$ respectively. $\sigma(\cdot) $ is the generalized Charbonnier penalty function, $\sigma = (x^2 + \epsilon ^ 2)^\gamma $, which is a differentiable, robust convex function \cite{Sun_IJCV}. We use the empirical values $\gamma = 0.45, \epsilon = 10^{-3}$ in this work.

\textbf{Regularizers.}
The form of the smoothness and divergence-free regularizer is
\begin{align}
    L_{s}({\bf{u}}^f, {\bf{u}}^b) + L_{div}({\bf{u}}^f, {\bf{u}}^b) = \sigma(\nabla {\bf{u}}^f) + \sigma(\nabla {\bf{u}}^b) + \notag \\ \sigma(\nabla \cdot {\bf{u}}^f) + \sigma(\nabla \cdot {\bf{u}}^b).
\end{align}
\label{eq:smoothness_loss}
The gradients and divergence of the velocity fields are approximated using a finite difference approach and computed by the convolution operator with appropriate filters. 

\subsection{Corrector}
\textbf{Temporal Loss.}
The training loss of the corrector is composed of two parts -- the temporal loss and the divergence loss:
\begin{equation}
    \begin{aligned}
    L_{C} = \underbrace{(\hat{\omega}_{t}- {\omega}_{t})}_{Temporal}+ \lambda_{d} L_{div},
    \end{aligned}
    \label{eq:total_loss_corrector}
\end{equation}
where $\omega = \nabla \times {\bf u}$ denotes the flow vorticity, which describes the tendency for fluid parcels to spin in the flow. The temporal loss is modelled by the difference between the warped vorticity field $\hat{\omega}_{t}$ from the last time level and the vorticity field ${\omega}_{t}$ computed using the corrected velocity field. As mentioned in section \ref{paragraph:transport}, warping the vorticity using the scheme \eqref{eq:warping_scheme_2} can be regarded as solving an advection-diffusion equation, such as the vorticity transport equation (shown in Appendix \ref{appendix:vorticity}). Therefore, reducing the temporal loss tries to enforce temporal consistency between current and previous steps.

%% file: experiment.tex
\section{Experiments}
\subsection{Datasets}\label{sec:dataset}
There are two kinds of dataset adopted in this work: synthetic and real world based. The PIV dataset is a synthetic dataset collected by \cite{Cai_1}. The dataset contains 15,050 particle image pairs with the originating flow field ground truth data obtained from computational fluid dynamics simulations. There are eight different types of flow contained in the dataset, including flow past a backward facing step (back-step) and past a cylinder, both at a variety of Reynolds numbers, DNS-turbulence, sea surface flow driven by a quasi-geostrophic model (SQG), etc.
For real world cases, we additionally use two datasets collected in a hydrodynamics laboratory in a shallow water flow past a cylinder configuration. To mimic the complex real world environment, we use imperfect markers in the form of foam and confetti as the visible tracers instead of high quality particles as would be typical in a professional PIV setting. The resulting images with imperfect and sparse tracer coverage are more challenging for fluid motion estimation, which helps extensively test the model's generalization ability.
\begin{figure*}
\centering
\includegraphics[width=1.0\textwidth]{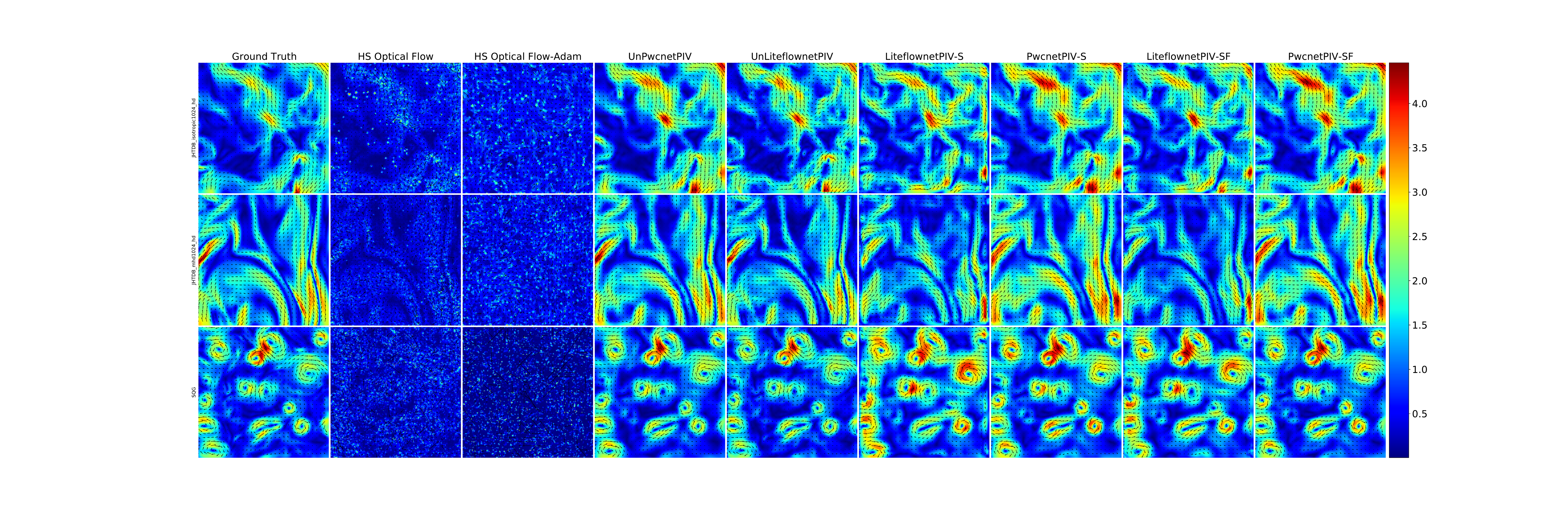}
\caption{Estimated flow for the samples in the PIV dataset. The color bar describes the magnitude of the displacements in pixels. }
\label{fig:PIV_result}
\end{figure*}
\subsection{Training}\label{sec:training} Due to the fact that the corrector can be combined with predictor in a non-intrusive way, we can first train the predictor without the corrector. The predictor is trained on the PIV dataset with 12,190 samples used for training and 1505 for testing. We train the model for 40,000 iterations with a batch size of four image pairs using the Adam optimiser. The learning rate is kept at $10^{-4}$. After the predictor is trained, the corrector is trained on time-resolved data with the predictor frozen. Here we use the flow type DNS-turbulence for corrector training. All experiments are conducted on a moderate level GPU Nvidia Tesla P100 16GB. We trained two unsupervised models, named UnPwcNet-PIV and UnLiteFlowNet-PIV to indicate that these are the unsupervised version of PwcNet and LiteFlowNet trained on the PIV dataset; along with four supervised models, LiteFlowNet-PIV-S, PwcNet-PIV-S, LiteFlowNet-PIV-SF and PwcNet-PIV-SF for comparison. The `S' denotes a supervised model, and `SF' represents a supervised model fine-tuned with double training iterations. Note the method named UnLiteFlowNet-PIV appeared in the author's previous work \citep{zhang2020unsupervised}, with the method proposed here further incorporating physics priors compared to the previous one.

\subsection{Results on Synthetic Dataset.}
\begin{table}[ht!]
\caption{Averaged end point error (AEPE) and averaged angular error (AAE) for the PIV dataset, the error unit is set to pixel per 100 pixels for easier comparison. From top to bottom, the first row shows the results of our own GPU implementation of the Horn–Schunck (HS) optical flow approach, the second row shows the results of the HS optical flow solved by an Adam optimizer. The results of the supervised learning models we trained are listed in the next four rows. The final two row shows results of our unsupervised methods introduced in this work. }
    \centering
    \resizebox{1.0\columnwidth}{!}{%
    \begin{tabular}{ccccccccccc}
    \toprule
     & 
    \multicolumn{2}{c}{Back-Step}&
    \multicolumn{2}{c}{Cylinder}&  
    \multicolumn{2}{c}{JHTDB}&
    \multicolumn{2}{c}{DNS}&
    \multicolumn{2}{c}{SQG}\\ 
    Methods& & & & &\multicolumn{2}{c}{channel} &\multicolumn{2}{c}{turbulence}  && \\
    \cmidrule(r){2-11}
    & AEPE& AAE& AEPE& AAE &AEPE& AAE  &AEPE& AAE &AEPE& AAE\\
    \midrule
    HS Optical Flow &221.1 & 78.7  & 88.7& 47.3  &39.4& 34.2 & 93.7& 58.7 &116.1& 70.0  \\
    HS Optical Flow-Adam &200.9 & 62.6  & 58.1&25.5 & 16.1& 13.5 & 57.7& 30.4 &70.3& 35.2\\
    \midrule
    LiteFlowNet-PIV-S& 16.4 & 8.3 & 15.5 & 6.8 & 22.4 & 18.9 & 37.6 & 21.3 & 44.4 & 23.5  \\
    PwcNet-PIV-S& 6.2 & 3.0 & 4.9 & 2.2 & 13.9 & 11.7 & 18.7 & 10.9 & 25.0 &  13.3  \\
    LiteFlowNet-PIV-SF& 13.7 & 6.9 & 13.0 & 5.6 & 19.2 & 16.2 & 30.8 & 17.6 & 36.0 & 18.9  \\
    PwcNet-PIV-SF& {\bf 4.7} & {\bf 2.4} & {\bf 4.1} & {\bf 1.8} & 12.9 & 10.8 & 16.0 & 9.4 & 22.1 & 11.8  \\
    \midrule
    \bf{UnLiteFlowNet-PIV}& 9.4 & 4.0 & 6.9 & 3.8 & {\bf 8.4} & {\bf 3.3} & {\bf 15.0} & {\bf 8.6} & {\bf 17.3} & {\bf 9.0}  \\
    \bf{UnPwcNet-PIV}& 8.2 & 3.9 & 7.1 & 3.9 & 13.4 & 11.3 & 21.5 & 12.8 & 25.2 & 13.5    \\
    \bottomrule
    \end{tabular}%
    }
\label{table:syn_results}
\end{table}

As shown in \ref{table:syn_results}, both the supervised and unsupervised learning methods outperform the HS optical flow baselines for most cases. UnPwcNet-PIV shows competitive performance compared to PwcNet-PIV-S. UnLiteFlowNet-PIV performs better than its related supervised model LiteFlowNet-PIV-S and even the fine-tuned (double the training iterations) model LiteFlowNet-PIV-SF. This suggests that the proposed unsupervised losses are reasonable and can be used as a substitute to ground truth data for training to some extent. Also, it indicates that the LiteFlowNet backbone is more suitable for the unsupervised learning method, while PwcNet can achieve better performance when trained in a supervised manner.



\begin{figure*}[ht!]
\centering
\includegraphics[width=1.0\textwidth]{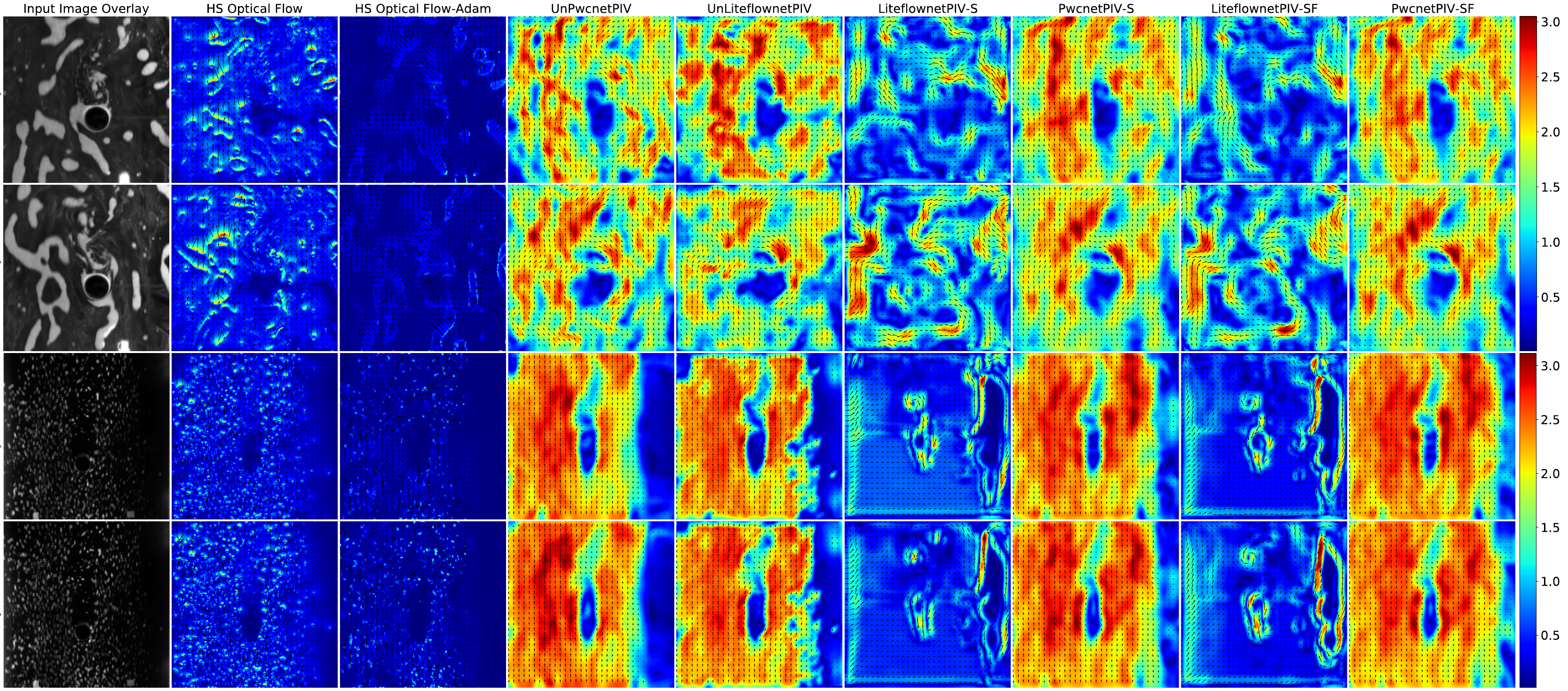}
\caption{Estimated flow on Fluid Foam (upper two rows) and Fluid Confetti (bottom two rows) dataset. The color bar describes the magnitude of the displacements in pixels.}
\label{fig:foam_confetti_result}
\end{figure*}

\textbf{Refinement by the Corrector.}
\begin{figure*}[ht!]
\centering
\minipage{0.3\textwidth}
\includegraphics[width=\linewidth]{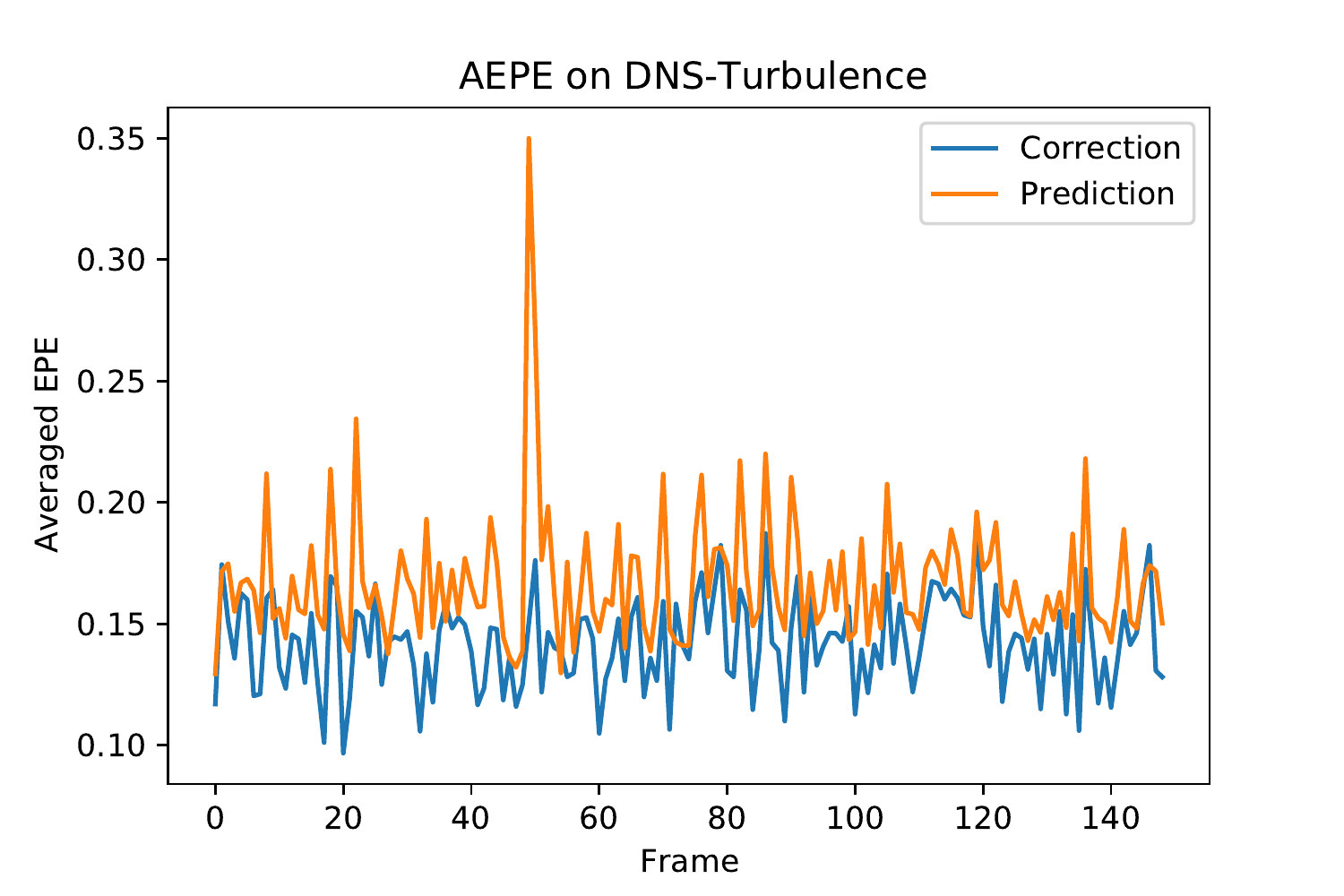}
\label{fig:refine_stat}
\endminipage\hfill
\minipage{0.7\textwidth}
\includegraphics[width=\linewidth]{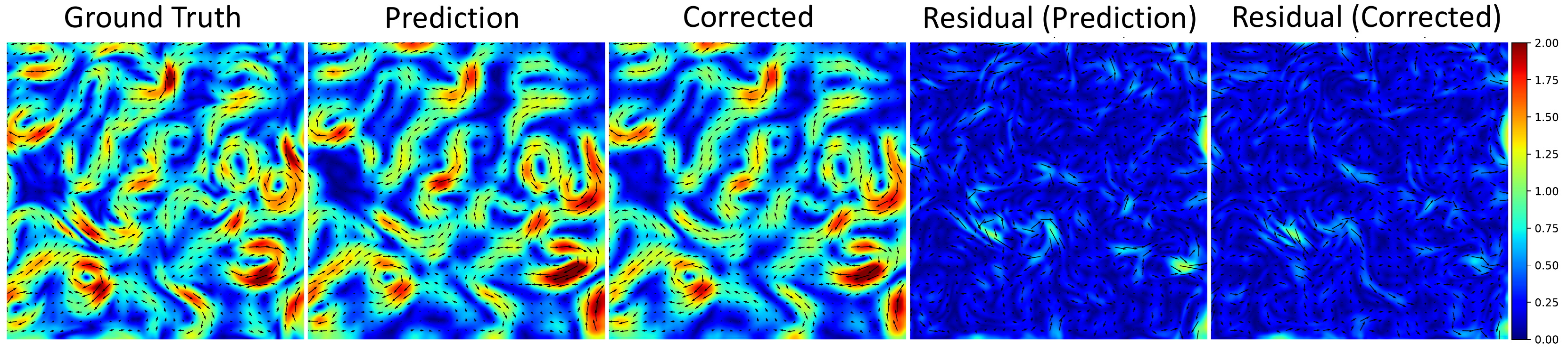}
\label{fig:refine_sample}
\endminipage\hfill
\caption{Comparison between prediction-only and the corrected velocities for the \textit{DNS-turbulence} data.}
\label{fig:refine}
\end{figure*}
Figure \ref{fig:refine} shows the abilities of the corrector for refining the prediction results. It can be observed that the corrector helps to reduce the average end point error and suppress the error fluctuation due to temporal inconsistency.

\subsection{Results on Real World Datasets}
\label{sec:real_world_case_analysis}

\textbf{Visual Comparisons.}
The results of the models when applied to the Fluid Foam and Fluid Confetti cases are shown in Figure \ref{fig:foam_confetti_result}. It can be observed that the HS optical flow and HS optical flow-Adam approaches struggle to give a clear estimation of the flow field. There are also obvious outliers around the edges of the foam or confetti. For supervised models, LiteflownetPIV-S and LiteflownetPIV-SF output estimations that conflict with the experimental setting, suggesting that the supervised models with the Liteflownet backbone overfit to the training dataset and show low generalization abilities. For the unsupervised models, the estimation results show reasonable flow structures while vorticity in the wake can also be observed. 

\textbf{Benchmark.}
Since there is no ground truth for real world datasets, here we used the best results (from our best tuning of the parameters) of the open source fluid motion estimation software PIVlab \cite{pivlab} (version 2.53) as a benchmark for our models. PIVlab is a popular Matlab toolbox, adopting correlation based, multi-pass, multi-grid window deformation techniques for fluid motion estimation, which can serve as a reliable baseline. We evaluate both the Fluid Foam and Fluid Confetti real use case datasets using PIVlab. For our models, we choose two unsupervised based models UnLiteFlowNet-PIV and UnPwcnet-PIV for comparisons. Note that the flow estimated velocity can also be interpreted as the displacement per unit time, so we use ``displacement'' in this paper for conciseness. 

\textbf{Quantitative Analysis.} We adopt three approaches to interpret the results on the real world dataset: wake analysis, displacement distribution analysis and image reconstruction. The results of these approaches are discussed in detail in the following paragraphs.  

\begin{figure}[b]
\includegraphics[width=1.0\columnwidth]{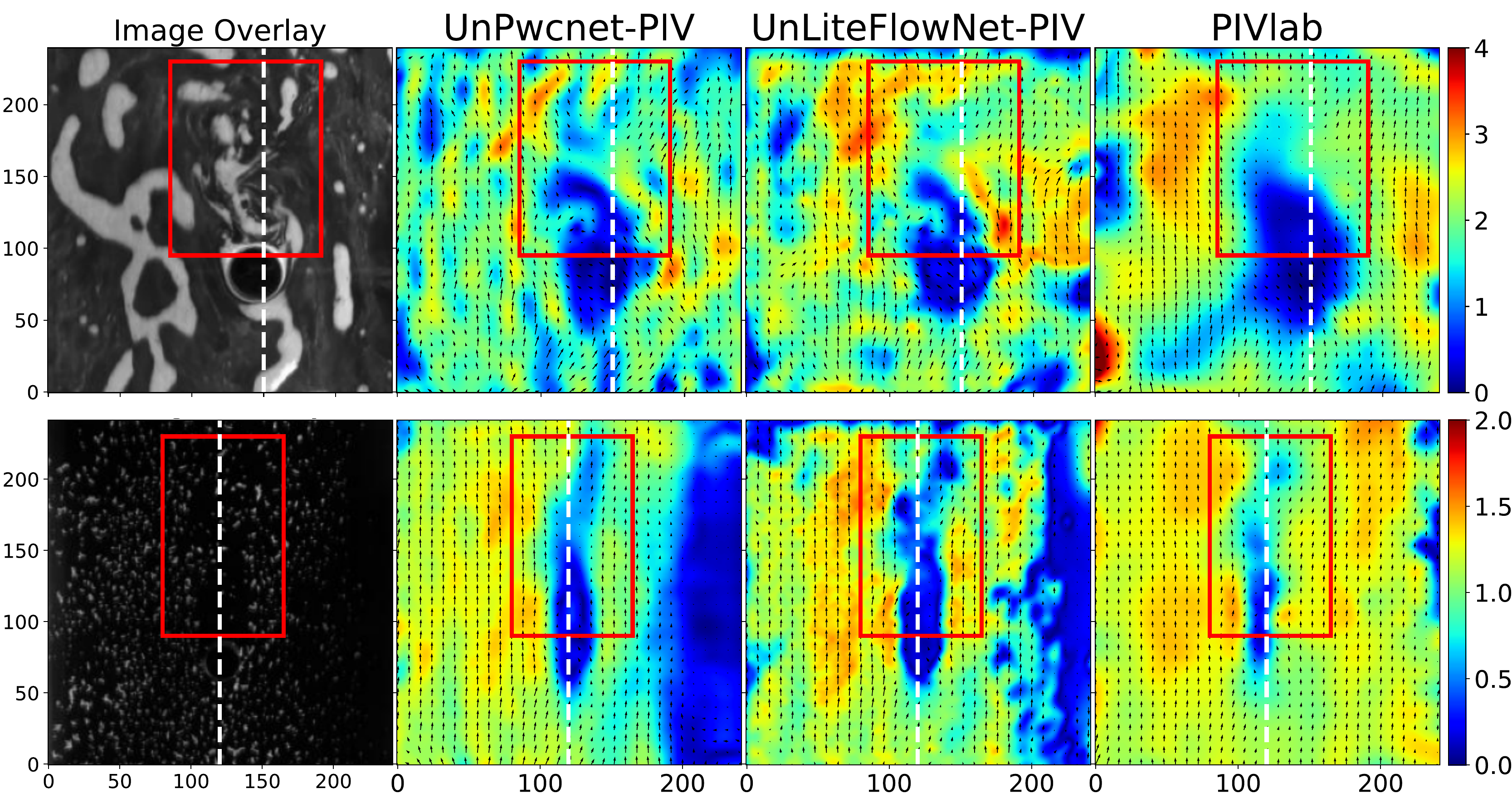} \\
\caption{\label{fig:visual}Visual results for Fluid Foam (upper row) and Fluid Confetti (bottom row) Dataset. The color bar describes the magnitude of the displacements in pixels.}
\end{figure}

\begin{figure}[ht!]
\centering
\includegraphics[width=0.49\textwidth]{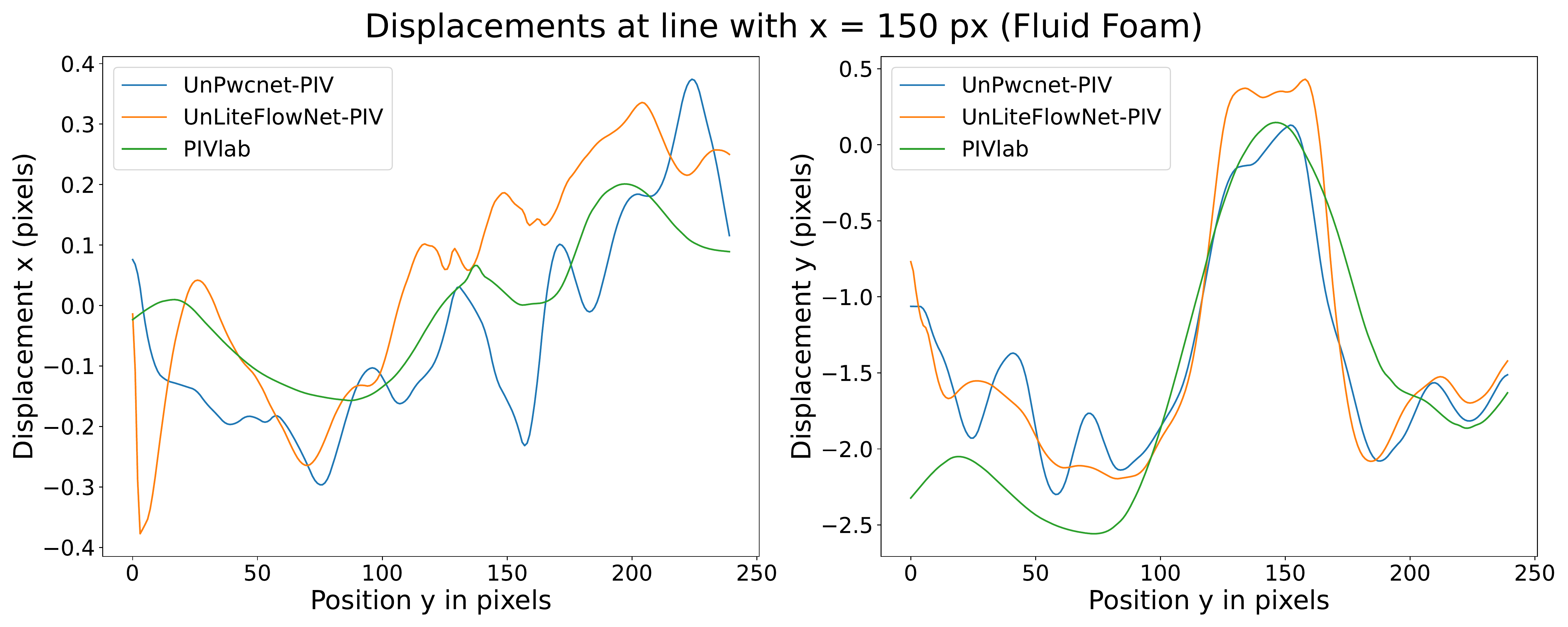}
\includegraphics[width=0.49\textwidth]{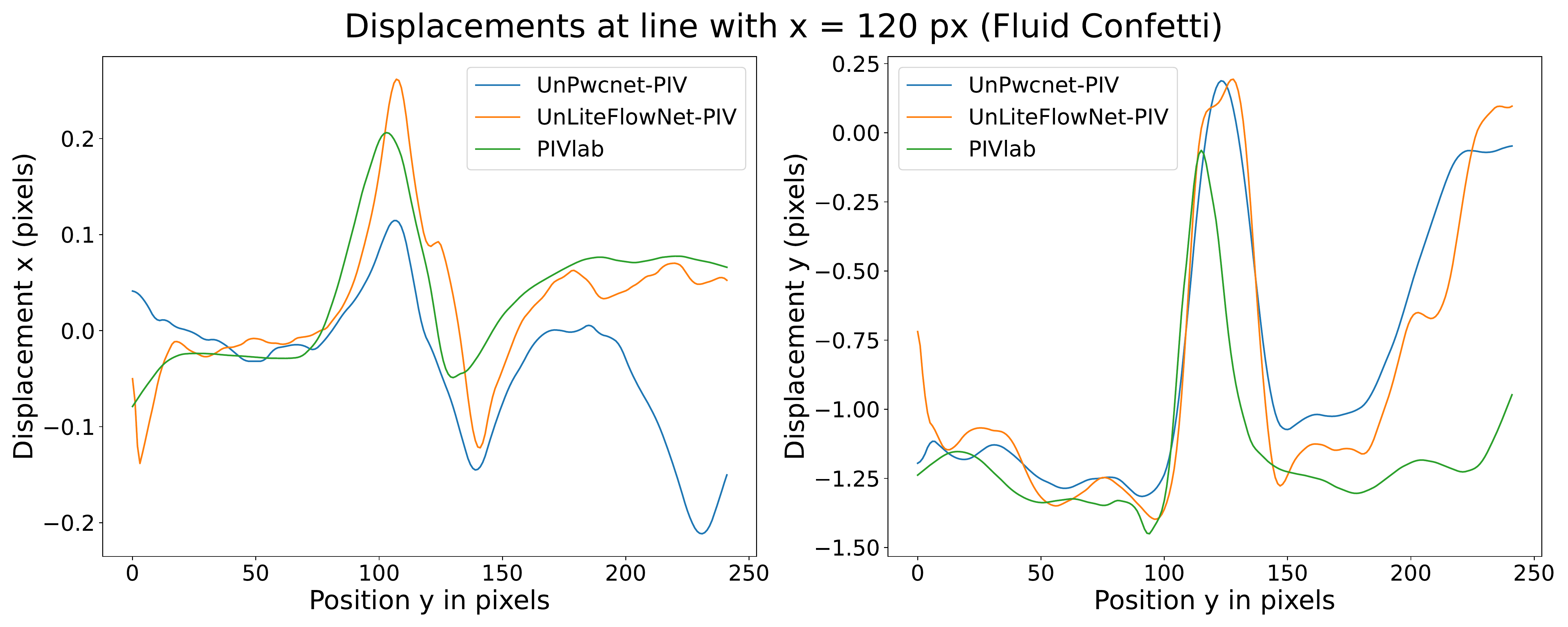}
\caption{\label{fig:wake} Estimated displacements (left column shows the displacement in the $x$ direction and the right column shows it in the $y$ direction) along the dashed white lines shown in Figure \ref{fig:visual}.}
\end{figure}

\textbf{Wake Analysis.} The hydrodynamics experiments performed here are for flow past a cylinder in a shallow water environment. The most interesting part of this problem is the wake dynamics (Karman vortex street). This occurs in the region downstream of the obstacle (i.e., the region in the red bounding box in Figure \ref{fig:visual}). Therefore, one way to interpret the results of these use cases is to note whether the method can capture wake dynamics properly.
To analyze the wake dynamics, we selected a line (shown by the vertical white dashed line in Figure \ref{fig:visual}) aligned with the position of the cylinder center, i.e. $x =$ 150 pixels for the Fluid Foam Dataset and $x =$ 120 pixels for the Fluid Confetti Dataset. We extracted the displacements for the $x$ and $y$ directions for all frames (195 for Fluid Foam and 175 for Fluid Confetti Dataset) along the line and compute the averaged displacements across frames. Figure \ref{fig:wake} shows that the displacement curves of our two models demonstrate similar tendency and magnitude to that obtained using PIVlab, which implies that our models output reasonable results and are able to achieve competitive performance compared with sophisticated, modern PIV software on these real use cases. In addition, the curves show periodic patterns after the back circle (at around $y =$ 100 pixels) of the cylinder, which is compatible to the dynamics expected of a Karman vortex street, indicating that the captured dynamics of all three methods are reasonable qualitatively. Note that due to its underlying methodology PIVlab can only output a sparse displacement field while the outputs of our methods are dense fields. To facilitate comparison, we interpolate the sparse field to a dense field using cubic interpolation. Therefore, the results of PIVlab are potentially over-smoothed, which may not properly describe the small scale dynamics such as vortices right after the cylinder. This may cause the gap between the curves obtained with our methods and PIVlab at the boundary regions.

\begin{figure}[ht!]
\centering
\includegraphics[width=0.49\textwidth]{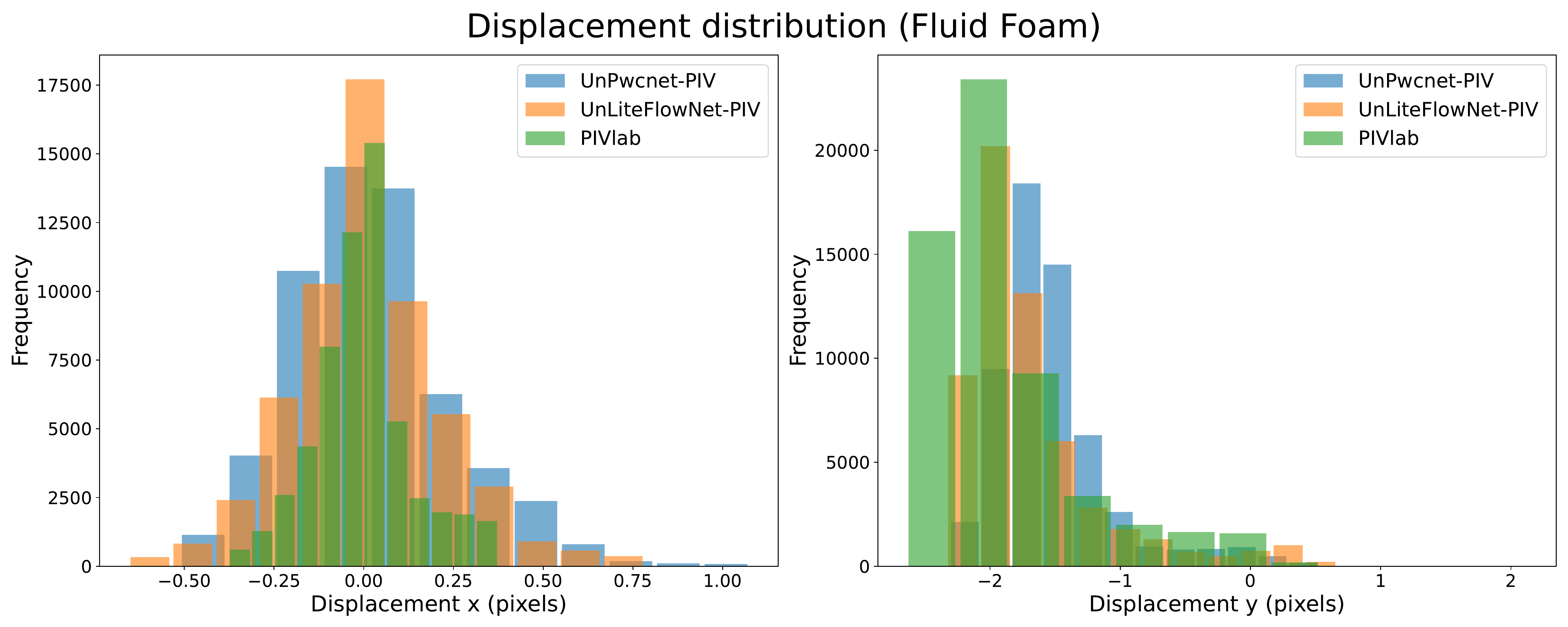}
\includegraphics[width=0.49\textwidth]{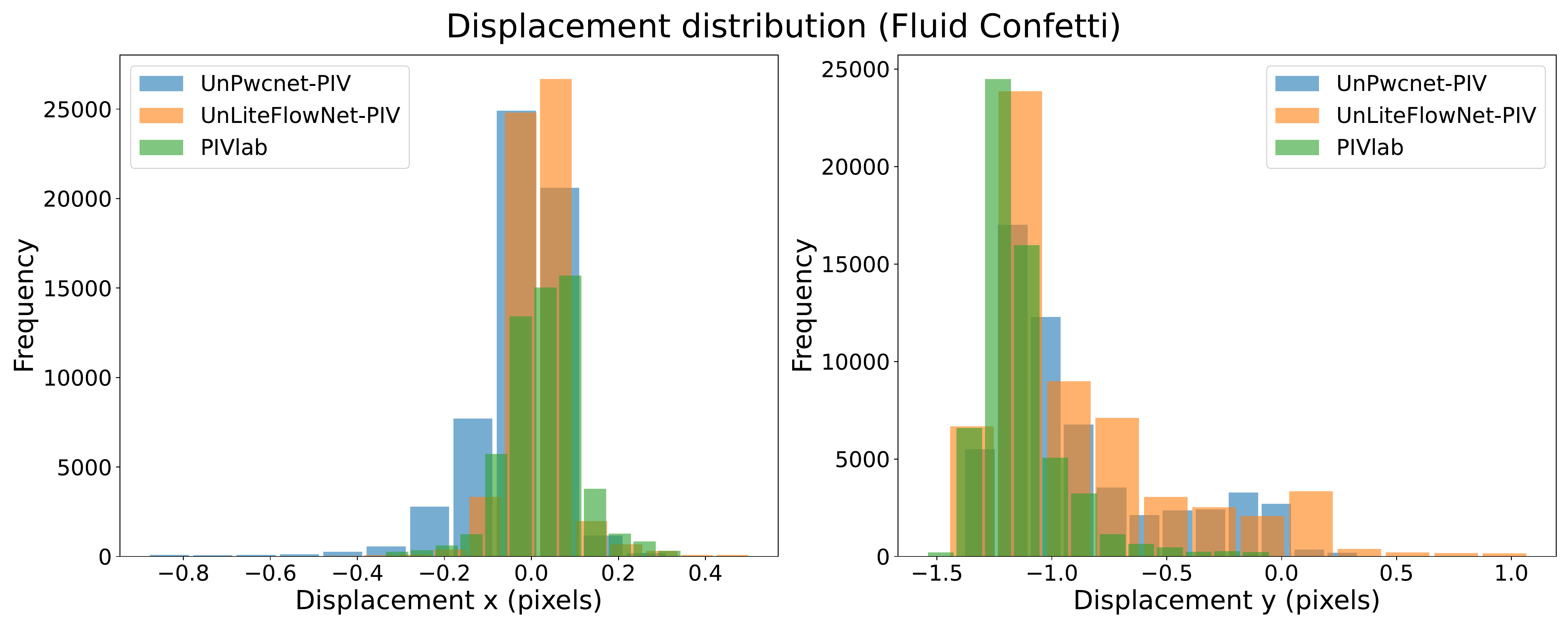}
\caption{\label{fig:distribution}Histograms of output displacements distribution for both Fluid Foam (upper row) and Fluid Confetti dataset (bottom row).}
\end{figure}

\textbf{Displacement Distribution} The statistic of the output displacement fields provide another measure which has been adopted by the PIV community for evaluating different algorithms \cite{piv4}, especially when there is no ground truth. Figure \ref{fig:distribution} shows the comparisons of output displacements distribution for both Fluid Foam (upper plot) and Fluid Confetti dataset (lower plot). It can be observed that the histograms of our methods and PIVlab largely overlap, indicating that output displacement distributions of these methods are similar.

\begin{figure}[ht!]
\includegraphics[width=1.0\columnwidth]{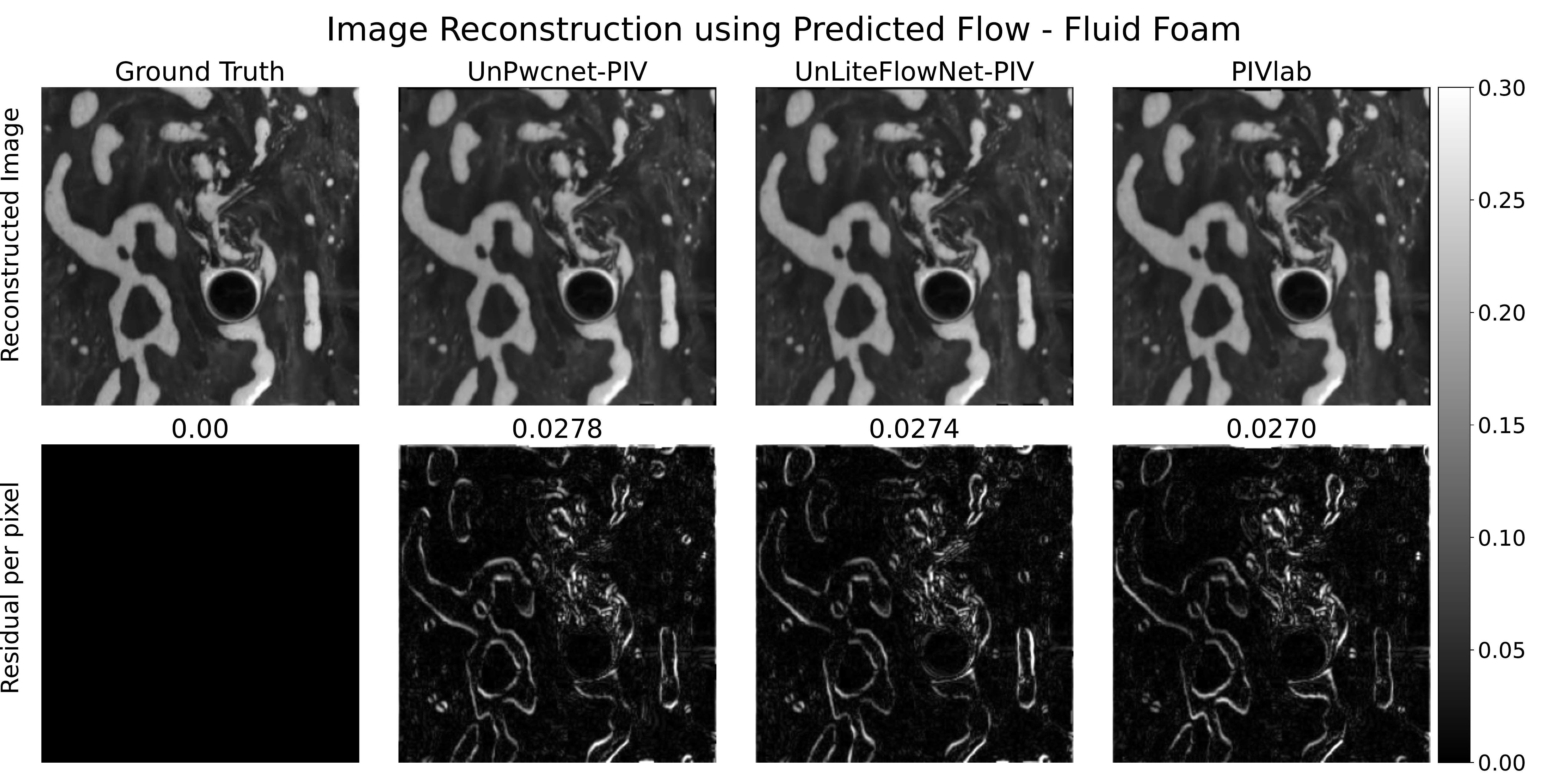}
\includegraphics[width=1.0\columnwidth]{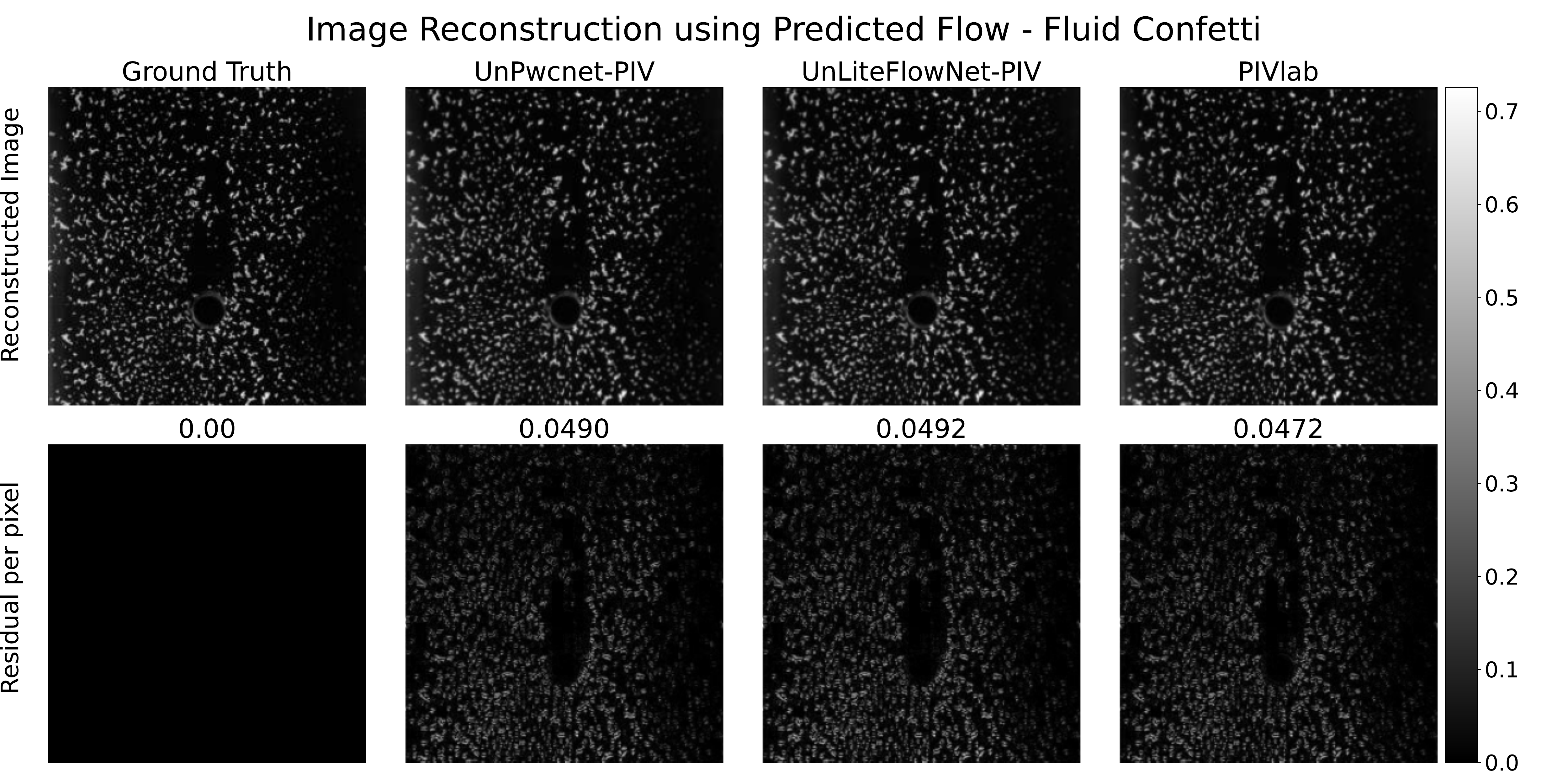}
\caption{\label{fig:reconstruction} Image reconstruction using the predicted flow on the Fluid Foam and Confetti datasets. The numbers above the second row of images indicate the residual per pixel between the reconstructed and the true images.}
\end{figure}

\textbf{Image Reconstruction.} Another way to verify the agreement of the results is to reconstruct future images using the estimated flow.  Figure \ref{fig:reconstruction} shows the future images reconstruction based on our methods and the PIVLab benchmark. It can be shown that the flow estimated by our methods can help reconstruct images close to ground truth visually, with competitive residual per pixel compared to PIVLab.

%% file: related_work.tex
\section{Related work} 
\textbf{Optical Flow Estimation.} Deep learning has shown its potential for optical flow estimation \citep{Dosovitskiy_flownet,Ilg_flowne2t,pwc,twhui}. These methods aim to estimate dense optical flow fields given an image pair using supervised trained deep neural network. There are also some efforts to investigate unsupervised \citep{Yu_back_to,Meister_unflow,what_matter} or self-supervised \citep{liu_2019,liu_aaai,what_matter} learning of optical flow estimation. All these methods are designed for pure optical flow methods, and mainly focus on the motion of rigid bodies. Moreover, there are works \cite{PIV-DCNN,Cai_1,Cai_2} adopting optical methods to fluid flow motion estimation. These methods are limited to supervised learning, while fluid data are difficult to collect and annotate in real world applications. In addition, fluid dynamics are not considered in these methods, thus the physical correctness of the estimation results are not guaranteed. In this work, we leverage physical knowledge and construct a corrector to refine the estimations for physical correctness.

\textbf{Learning to Solve PDEs.}
Solving Partial Differential Equations (PDEs) numerically is often slow and inefficient. To improve efficiency, several attempts have been conducted to
approximate the solution and response function of PDEs using neural networks \citep{PINN,hidden_physics,fourier_pde}. The connections between numerical schemes for solving PDEs and residual neural network have also been investigated \citep{neural_ode,bridging,E_dynamic,RK} for predicting dynamical systems. Also, some studies show that partial derivatives can be approximated with convolutions \citep{pde_net,pde_net2}, which can be used to discretize a broad class of PDEs. In this work, we leverage this idea and use convolutions to model the fluid dynamics.

\textbf{Learning to Enhance Fluid Simulation.} Various studies have investigated the application of deep learning techniques to enhance fluid simulation, in terms of either efficiency or accuracy. Several studies have sought to accelerate simulation using deep neural networks. For solving the incompressible Euler equations, convolution networks are trained for divergence-free corrections \citep{accelerate_divergence}. In turbulent flows modelling \citep{google_cfd}, deep neural networks are applied to accelerate both direct numerical simulation and large eddy simulation of turbulent flows. To reduce numerical errors, a differentiable physics network is proposed which can interact with iterative PDE solvers   \cite{solver_in_loop}. Existing studies have also explored the abilities to learn to simulate complex physics simulations using of graph neural networks, in both Eulerian \citep{learn_to_simulate_mesh} and Lagrangian \citep{learn_to_simulate_particle} perspectives. Graph neural networks are also utilized to process unstructured mesh data and speed-up fluid flow prediction \citep{cmu_graph} in computational fluid dynamics. However, these methods focus on simulation only; they do not support the processing of observations at each time step, which is required in fluid motion estimation scenarios.

%% file: conclusion.tex
\section{Conclusion}
\textbf{Summary.} We present here an unsupervised learning approach for solving the problem of fluid flow motion estimation. The proposed approach shows significant promise and potential advantages for fluid flow estimation. It yields competitive results when compared to existing supervised learning based methods, and even outperforms them for some difficult flow cases. Furthermore, the unsupervised learning approach shows robust generalization ability when dealing with complex real-world flow scenarios, though trained purely on synthetic data. In addition, the proposed physical corrector is able to refine the estimates obtained from the predictor by incorporating temporal information and fluid knowledge.

\textbf{Limitation and Future Work.}\label{sec:limitation} There are several limitations of the approach presented in this paper. First, there are only limited training and testing fluid observation data available, which is not enough for extensive test of the corrector. To address this, we aim to collect and generate more time-resolved data of different flow types for further investigation. Second, the initial and boundary conditions are not known for the physical model in the corrector. Third, the current model relies on observations at every time step. We plan to further investigate the prediction-correction scheme for forecasting tasks. This extension will require network architectures such as recurrent neural networks (RNNs) to better handle temporal information. Finally, we would like to explore more learning based methods inspired by numerical simulation, to bridge the communities between machine learning and scientific computing.

\textbf{Acknowledgements.}
This work is sponsored by the Chinese Scholarship Council and Imperial College London (under a pump priming research award from the Energy Futures Lab, Data Science Institute and the Gratham Institute). JT wishes to acknowledge PhD research funding from the  Government of Botswana.

%% file: appendix.tex
\newpage
\appendix
\onecolumn


\section{Fluid dynamics}
Here we introduce the key methods and equations of fluid dynamics in this work.
\paragraph{Stokes Equation.}
The Stokes equation
\begin{equation}
    \begin{aligned}
     - \mu \nabla^2 {\bf {u}} + \nabla p = {\bf{f}},
    \end{aligned}
    \label{eq:stokes_equation}
\end{equation}
where $\mu$ is the dynamic viscosity, is an approximation/simplification to the Navier–Stokes momentum equation obtained by omitting the inertial part. It is usually used to model fluid flow with a low Reynolds number, where advective inertial forces are small compared with viscous forces. In this work, we demonstrate that the output of the motion predictor can be interpreted as the solution of a Stokes-like equation.

\paragraph{Vorticity Transport Equation.}
\label{appendix:vorticity}
The vorticity is a pseudovector field (scalar field in 2D) that describes the tendency for fluid parcels to spin in the flow. Here we consider the incompressible vorticity transport equation that takes the form 
\begin{equation}
    \begin{aligned}
     \omega_t + {\bf u} \cdot \nabla{\omega} = \nu \nabla^2{{\omega}},
    \end{aligned}
    \label{eq:vorticity_equation}
\end{equation}
where $\omega = \nabla \times {\bf u}$ denotes the vorticity, and $\nu$ is the kinematic viscosity. This is closely related to the full incompressible Navier-Stokes equation for homogeneous flow, describing the evolution of the fluid’s vorticity over time. By using the vorticity transport equation, we can evolve the vorticity field while avoiding the need to solve the coupled pressure field that appears in the full Navier-Stokes system.

\paragraph{Chorin's Method.}
\label{chorin_method}
Chorin’s projection method can be considered as an operator splitting based approach. The idea is first to compute a tentative velocity ${\bf u}_t^*$ by neglecting the pressure in the Navier-Stokes momentum equation, i.e. by solving 
\begin{equation}
    \begin{aligned}
      \frac{{\bf u}_t^* - {\bf u}_{t-1}}{\Delta t} = - {\bf u}_{t-1} \cdot \nabla{{\bf u}_{t-1}} + \nu \nabla^2{{\bf u}_{t-1}},
    \end{aligned}
    \label{eq:chorin_1_fd}
\end{equation}

and then projecting the obtained velocity ${\bf u}_t^*$ onto the space of divergence free vector fields using the update

\begin{equation}
    \begin{aligned}
      \frac{{\bf u}_t - {\bf u}_t^*}{\Delta t} = - \frac{1}{\rho} \nabla{p_{t}}.
    \end{aligned}
    \label{eq:chorin_2_fd}
\end{equation}

Note that if we add up both sides of Equation \eqref{eq:chorin_1_fd} and \eqref{eq:chorin_2_fd}, the incompressiable Navier-Stokes momentum equation is recovered with the tentative velocity ${\bf u}_t^*$ cancelled.

According to Helmholtz decomposition, the velocity ${\bf u}^*_t$ can be decomposed into a divergence-free (solenoidal) part and an irrotational part:
\begin{equation}
    \begin{aligned}
      {\bf u}^*_t = {\bf u}_{\text{sol}} + {\bf u}_{\text{irrot}},
    \end{aligned}
    \label{eq:h_decomposition}
\end{equation}
where $\nabla \cdot {\bf u}_{\text{sol}} = 0$. The ${\bf u}_{\text{sol}}$ is the velocity ${\bf u}_t$ we would like to solve. Therefore, by taking the divergence of both sides of Equation \eqref{eq:chorin_2_fd}, we obtain the equation 
\begin{equation}
    \begin{aligned}
          \frac{\rho }{\Delta t}\nabla \cdot {\bf u}_t^* = \nabla^2{p_{t}},
    \end{aligned}
    \label{eq:chorin_3_fd}
\end{equation}
which is a Poisson equation for the unknown pressure $p_t$. Following solution of this equation, the updated velocity can be computed as 
\begin{equation}
    \begin{aligned}
      {\bf u}_t  = {\bf u}_t^* - \frac{\Delta t}{\rho} \nabla{p_{t}}.
    \end{aligned}
    \label{eq:chorin_4_fd}
\end{equation}

\paragraph{Warping Scheme and Advection-Diffusion Equation.}
\label{warping_proof}
\paragraph{Theorem 1.} For any initial condition $I_0 \in L^1(\mathbb{R}^2)$ with $I_0(\pm\infty)=0$, there exists a unique global solution $I(x, t)$ to the advection-diffusion equation
\begin{equation}
    \begin{aligned}
     I({\bf x}, t) = \int_{\mathbb{R}^2} k({\bf x}-{\bf u}, {\bf y})I_0({\bf y})d{\bf y},
    \end{aligned}
    \label{eq:warping_scheme_1}
\end{equation}
where $k(a, b) = \frac{1}{4\pi Dt}e^{-\frac{1}{4Dt}\left\| a - b \right\|^2}$ is a Gaussian distribution density with mean ${\bf x}-{\bf u}$ and variance $2Dt$, ${\bf u}$ is the velocity field. Equation \eqref{eq:warping_scheme_1} indicates that the scalar field $I (x, t)$ can be computed via a convolution between a Gaussian kernel and the initial condition $I_0$. The full proof of the theorem can be found in \citet{dl_physics_process}.

\section{Euler-Lagrangian Equation Derivation}
\label{el_proof}
Here we show the derivation of the Euler-Lagrange equation of formulation (6). Given ${\bf{u}} = (u, v)$, denoting $f_i = \frac{\partial f}{\partial x_i}, f_{ij} = \frac{\partial f}{\partial x_i \partial x_j}$, we rewrite Equation (6) as:
\begin{equation}
    \begin{aligned}
        \mathcal{F}(u, v) =& \int_\Omega \mathcal{L}(u, v, u_x, u_y, v_x, v_y) \ d{\bf{x}} \\ 
        =& \int_\Omega (I_t + uI_x +vI_y) + \mu (u_x^2 + u_y^2 + v_x^2 + v_y^2) \\ &+ p (u_x + v_y) \ d{\bf{x}}.
    \end{aligned}
\end{equation}
According to the definition of Euler-Lagrange equation for several functions of several variables with single derivative, we can get the following system of equations for $u$ and $v$:
\begin{equation}
    \begin{aligned}
    \begin{cases}
        \frac{\partial \mathcal{L}}{\partial u} - \frac{\partial}{\partial x}(\frac{\partial \mathcal{L}}{\partial u_x}) - \frac{\partial}{\partial y}(\frac{\partial \mathcal{L}}{\partial u_y}) = 0, \\ 
        \frac{\partial \mathcal{L}}{\partial v} - \frac{\partial}{\partial x}(\frac{\partial \mathcal{L}}{\partial v_x}) - \frac{\partial}{\partial y}(\frac{\partial \mathcal{L}}{\partial v_y}) = 0. 
    \end{cases}
    \end{aligned}
\end{equation}
By simplifying the equations above, we can get
\begin{equation}
    \begin{aligned}
    \begin{cases}
        2 \mu (u_{xx} + u_{yy}) + p_x = I_x, \\ 
        2 \mu (v_{xx} + v_{yy}) + p_y = I_y. 
    \end{cases}
    \end{aligned}
\end{equation}
Rewriting the system of equations above, we can get
\begin{equation}
    \begin{aligned}
        -2\mu \nabla ^ 2 {\bf u} + \nabla (-p) = \nabla (-I), 
    \end{aligned}
\end{equation}
which has the same form as the Stokes equation, where $\mu$ is the dynamic viscosity constant, $\nabla (-p)$ and $\nabla (-I)$ are the pressure gradient term and the external force term, respectively. Considering that $\mu$ is a constant we can omit the constant 2 in front of it. For the force terms, the minus sign only denotes the direction of the forces and thus they are omitted in Equation \eqref{eq:stokes_equation_predictor} of the main paper for clarity.

\section{Further Generalization Tests}
\begin{figure}[ht!]
\centering
\includegraphics[width=1.0\textwidth]{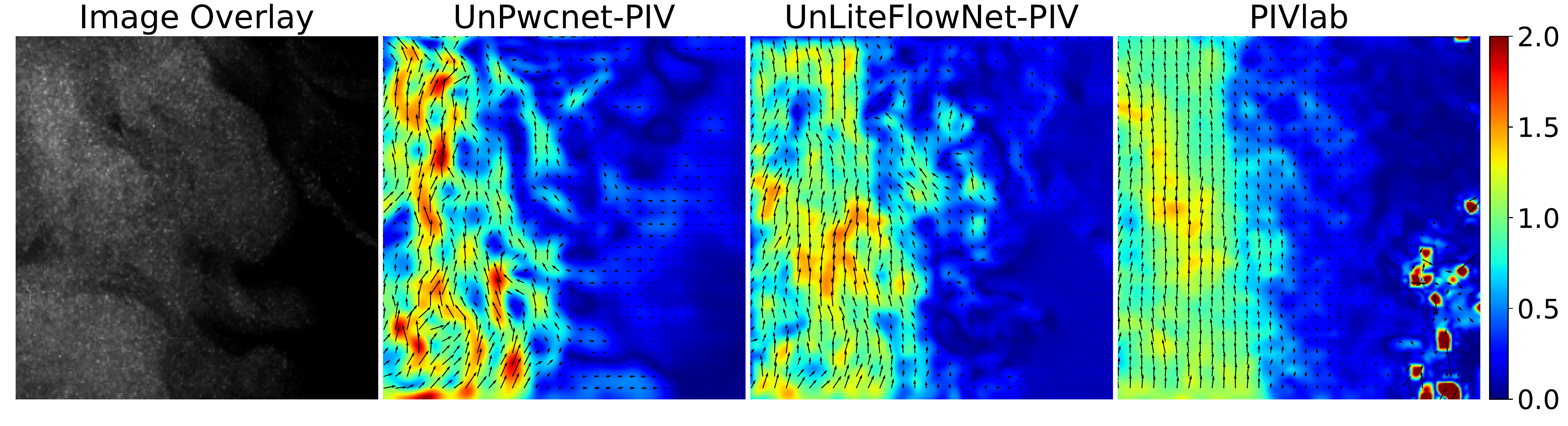}
\includegraphics[width=1.0\textwidth]{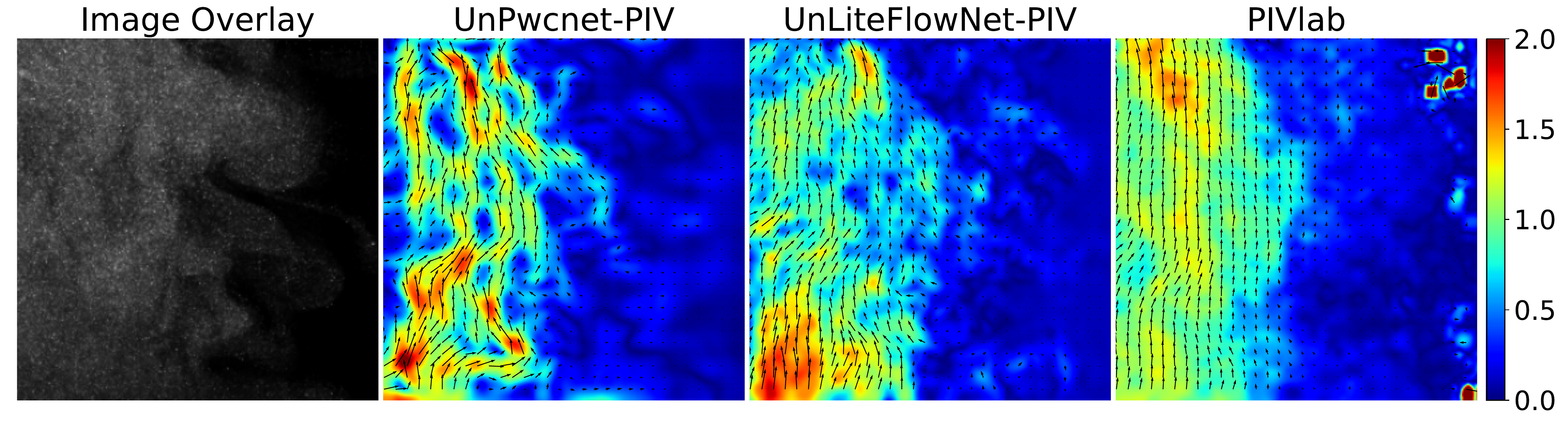}
\includegraphics[width=1.0\textwidth]{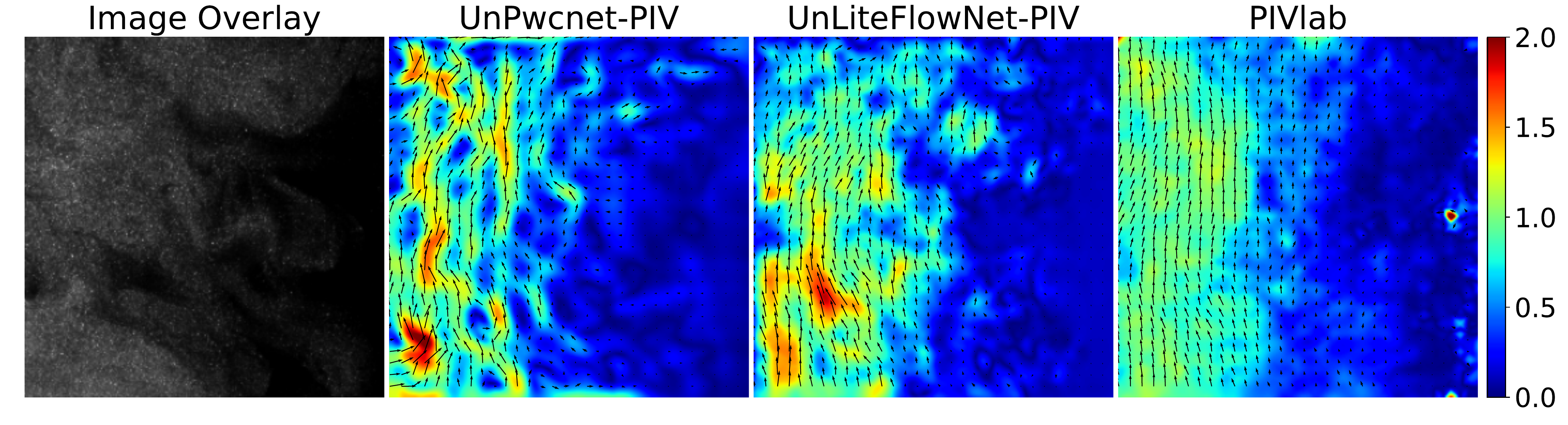}
\caption{\label{fig:jet_flow}Extra real use case ``Jet Flow'' from 3th PIV challenge \url{https://www.pivchallenge.org/}. The figure shows that our methods can capture the dynamics properly with more small scale detail comparing to the over-smoothed PIVlab results.}
\end{figure}

\begin{figure}[ht!]
\centering
\includegraphics[width=1.0\textwidth]{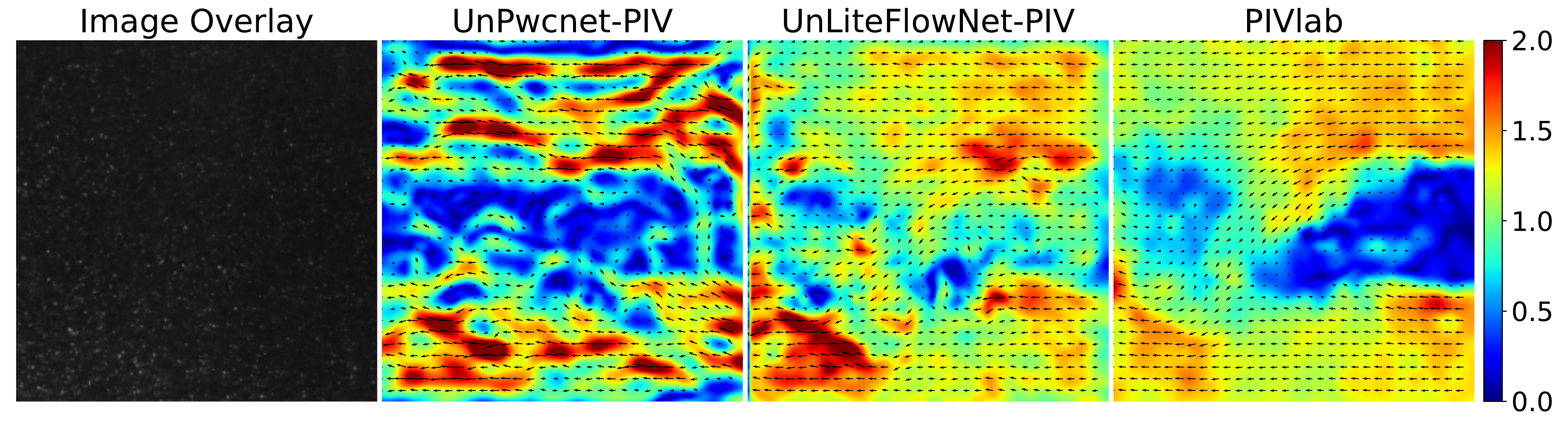}
\includegraphics[width=1.0\textwidth]{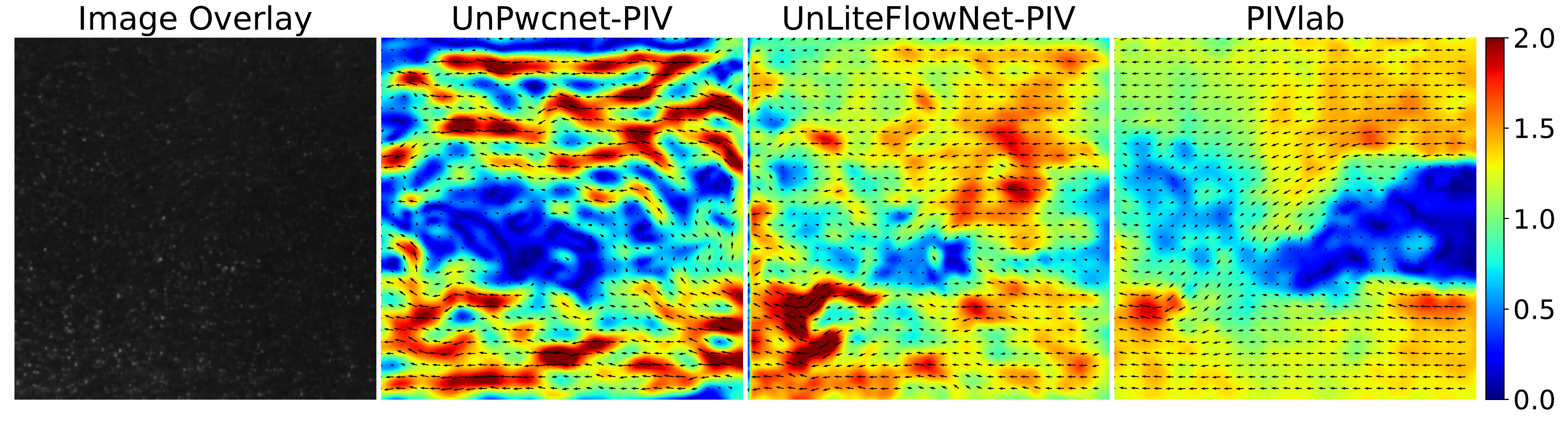}
\includegraphics[width=1.0\textwidth]{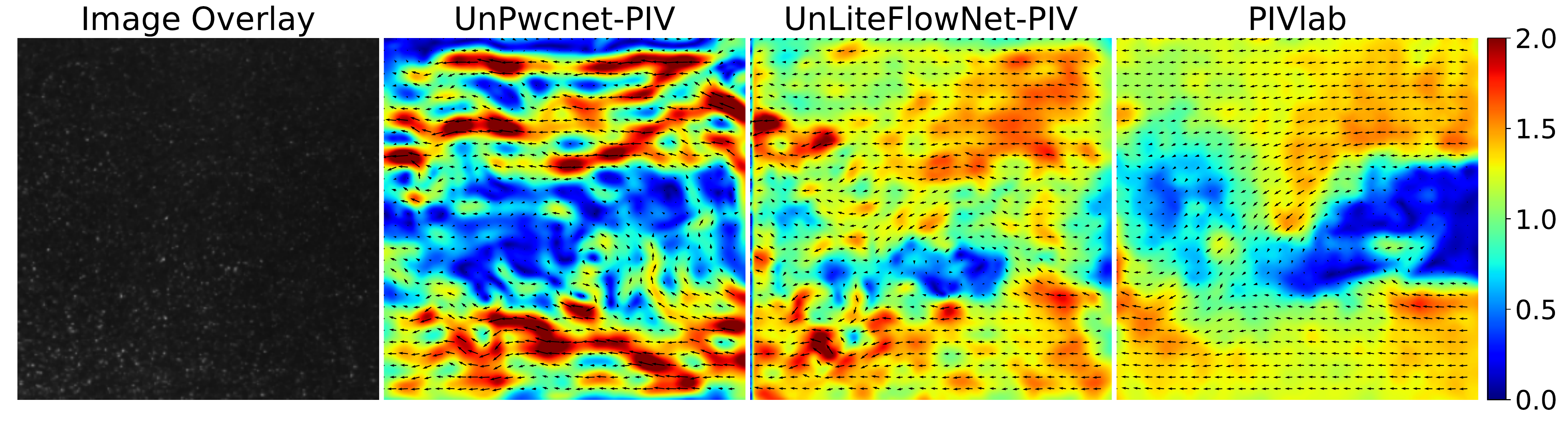}
\caption{\label{fig:karman}Extra real use case ``Karman'' from PIVlab. It is observed that the model UnLiteFlowNet-PIV can still capture the wake after the obstacle, although the UnPwcnet-PIV outputs noisy results.}
\end{figure}

 Although the Fluid foam and Fluid confetti dataset mentioned in this paper have already served to test the generalization abilities of the models, we adopt additional test cases from the Particle Image Velocimetry community. We consider two examples: ``Jet Flow'' (shown in Figure \ref{fig:jet_flow}) and ``Karman'' (shown in Figure \ref{fig:karman}). 

\section{Dataset}

\paragraph{PIV Dataset.}
There are eight different types of flow in the PIV dataset. The types ``Back-step'' and ``Cylinder'' also contain data for different Reynolds numbers. The detailed descriptions for each type are shown in Table \ref{table:piv_dataset}.

\begin{table}[ht]
\caption{Detailed description of the PIV dataset considered, from \cite{Cai_1}. $dx$ refers to the particle displacements considered between two image frames in units of number of pixels. Re refers to the Reynolds numbers considered. `JHTDB' implies that the data was taken from the Johns Hopkins turbulence databases \cite{JHTDB}. Refer to \cite{Cai_1} for further details.}\label{table:piv_dataset}
\begin{center}
\begin{tabular}{cccc}
\toprule
Type Name& Description & Condition& Quantity\\
\toprule
Uniform & Uniform flow & $|dx| \in [0, 5]$ & 1000   \\
\midrule
&Flow past a backward facing step
   & Re = 800  & 600  \\
 Back-step&  & Re = 1000  & 600  \\
 &  & Re = 1200  & 1000 \\ 
 &  & Re = 1500  & 1000 \\  
\midrule
&Flow past a circular cylinder & Re = 40  & 50  \\
 &  & Re = 150  & 500  \\
 Cylinder&  & Re = 200  & 500 \\ 
 &  & Re = 300  & 500 \\
 &  & Re = 400  & 500 \\
\midrule
DNS-turbulence &  Homogeneous and    & -  & 2000  \\
 &isotropic turbulent flow &  &  \\
\midrule
SQG & Sea surface flow     & -  & 1500  \\
 &driven by SQG model &  &  \\
\midrule
Channel flow &  Channel flow    & -  & 1600  \\
 &provided by JHTDB &  &  \\
\midrule
JHTDB-mhd1024  &  Forced MHD turbulence     & -  & 800  \\
 &provided by JHTDB  &  &  \\
\midrule
JHTDB-isotropic1024 &  Forced isotropic turbulence   & -  & 2000  \\
 &provided by JHTDB &  &  \\
\bottomrule
\end{tabular}
\end{center}
\end{table}

\begin{figure}[ht]
\begin{center}
\includegraphics[width=1.0\textwidth]{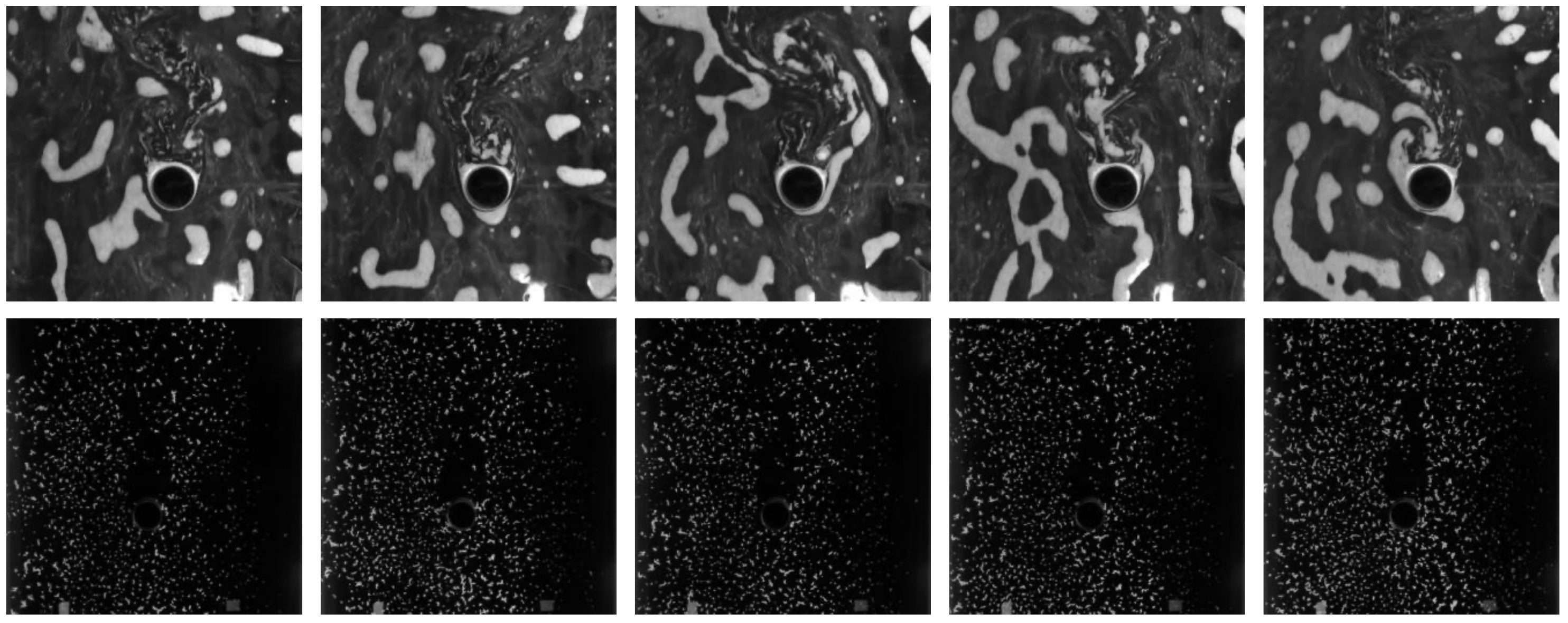}
\caption{Samples of Fluid Foam (upper row) and Fluid Confetti (bottom row) dataset. The white foam/confetti in the image serve as the visible tracers. } \label{fig:real_world_dataset}
\end{center}
\end{figure}

\paragraph{Fluid Foam Dataset.}
The setting for Fluid Foam can be summarized as follows: cylinder diameter = 50mm, frame rate = 60fps, water depth = 90mm, 1px = 0.0002174m. A sample of the collected images is shown in Figure \ref{fig:real_world_dataset}.


\paragraph{Fluid Confetti Dataset.}
Fluid Confetti is another dataset that we collected using a similar setting as for the Fluid Foam case. The difference is that the visible tracers are small pieces of white confetti (small squares of paper around 2mm in width), and the lighting has also been set up so the tracers could be clearly observed in the images. For the camera, a Hero7 camera with 4k resolution was set up at an angle and a 2.7k resolution Hero5 camera was set up vertically. The camera takes photos with an interval 1/60 second. 

